\documentclass[11pt]{article}
\usepackage{coling2020}
\usepackage{times}
\pdfoutput=1

\usepackage{url}
\usepackage{latexsym}
\usepackage{soul}
\usepackage{graphicx}
\usepackage{amsmath}
\usepackage{amsthm}
\usepackage{booktabs}
\usepackage{algorithm}
\usepackage{algorithmic}
\usepackage{multirow}
\usepackage{graphicx}
\usepackage{amssymb}
\usepackage{amsfonts}
\usepackage[hidelinks]{hyperref}
\usepackage{fancyvrb}
\usepackage{todonotes}

\colingfinalcopy 

\newcommand{\citep}[1]{\cite{#1}}

\title{A survey of embedding models of entities and relationships \\ for knowledge graph completion}

\author{Dat Quoc Nguyen \\
  VinAI Research, Vietnam \\
  {\tt v.datnq9@vinai.io} }

\date{}

\begin{document}
\maketitle

\begin{abstract}
\noindent Knowledge graphs (KGs) of real-world facts about entities and their relationships are useful resources for a variety of natural language processing tasks. However, because knowledge graphs are typically incomplete, it is useful to perform \emph{knowledge graph completion} or \textit{link prediction}, i.e. predict whether a relationship not in the knowledge graph is likely to be true. This paper serves as a comprehensive  survey of embedding models of entities and relationships for knowledge graph completion, summarizing up-to-date experimental results on standard benchmark datasets  and pointing out potential future research directions.


\end{abstract}

\section{Introduction}

\begin{figure}[t]
    \centering
    \begin{minipage}{0.475\textwidth}
        \centering
\includegraphics[width=8cm]{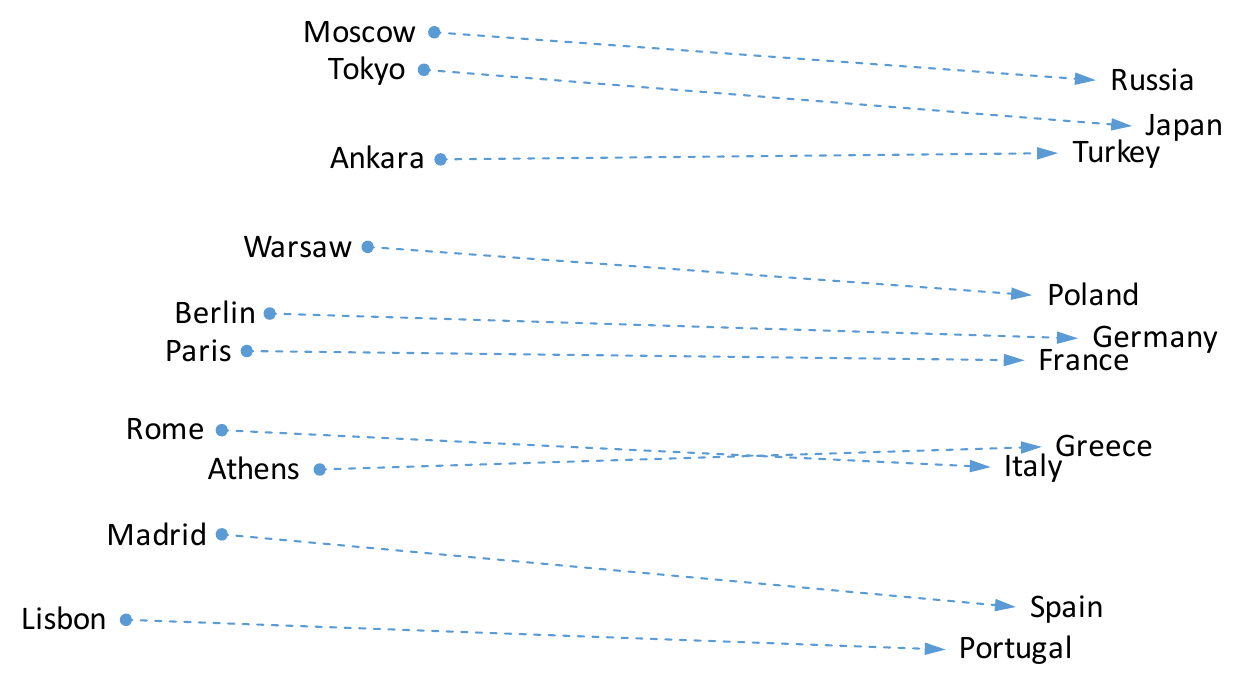} 
\caption[Two-dimensional  projection of vectors of countries and their capitals.]{Two-dimensional  projection of   vectors of countries and their capitals. This figure is drawn based on \newcite{MikolovNIPS2013}.}
\label{tab:countrycapital}
    \end{minipage}\hfill
    \begin{minipage}{0.475\textwidth}
        \centering
\includegraphics[width=4.4cm]{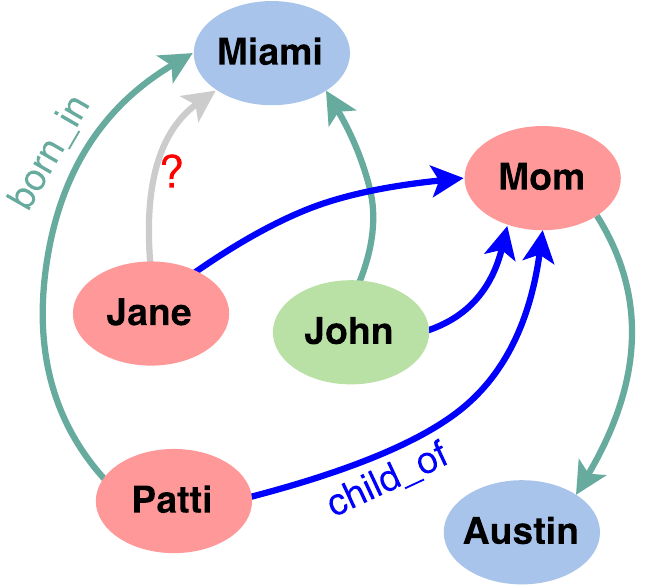} 
\caption[An illustration of (incomplete) knowledge base, with 4 person entities, 2 place entities, 2 relation types and total 6 triple facts.]{An illustration of (incomplete) knowledge base, with 4 person entities, 2 place entities, 2 relation types and total 6 triple facts. This figure is drawn based on  \protect{\newcite{westonembedding}}.}
\label{fig:kbexample}
    \end{minipage}
\end{figure}

Let us revisit the classic Word2Vec example  of a ``royal''   relationship   between ``$\mathsf{king}$'' and ``$\mathsf{man}$'', and  between ``$\mathsf{queen}$'' and ``$\mathsf{woman}$''. As illustrated in this example: $\boldsymbol{v}_{king} - \boldsymbol{v}_{man} \approx \boldsymbol{v}_{queen} - \boldsymbol{v}_{woman}$, word  vectors learned from a large   corpus can model relational similarities or linguistic regularities between pairs of words  as translations in the projected vector space  \citep{MikolovNIPS2013,Pennington14}. Figure \ref{tab:countrycapital} shows another example of  a relational similarity between word pairs of countries and capital cities:

\setlength{\abovedisplayskip}{-5pt}
\begin{eqnarray*}
\boldsymbol{v}_{Japan} - \boldsymbol{v}_{Tokyo} &\approx& \boldsymbol{v}_{Germany} - \boldsymbol{v}_{Berlin}\\
\boldsymbol{v}_{Germany} - \boldsymbol{v}_{Berlin} &\approx& \boldsymbol{v}_{Portugal} - \boldsymbol{v}_{Lisbon}
\end{eqnarray*}

Assume that we consider the country and capital pairs in Figure \ref{tab:countrycapital} to be pairs of entities rather than word types. That is, we now represent country and capital entities by low-dimensional and dense vectors.
The relational similarity  between word pairs is presumably to capture a ``$\mathsf{is\_capital\_of}$'' relationship between country and capital entities. Also, we represent this relationship by a translation vector $\boldsymbol{v}_{{is\_capital\_of}}$ in the entity  vector space. Thus, we expect: 

\begin{eqnarray*}
\boldsymbol{v}_{Tokyo} + \boldsymbol{v}_{{is\_capital\_of}} - \boldsymbol{v}_{Japan} &\approx &  \boldsymbol{0} \\
\boldsymbol{v}_{Berlin} + \boldsymbol{v}_{{is\_capital\_of}} - \boldsymbol{v}_{Germany}  &\approx &  \boldsymbol{0} \\
\boldsymbol{v}_{Lisbon} + \boldsymbol{v}_{{is\_capital\_of}} - \boldsymbol{v}_{Portugal}  &\approx & \boldsymbol{0} 
\end{eqnarray*}

\noindent  This intuition inspired the TransE  model---a well-known embedding model  for KG completion or link prediction in KGs   \citep{NIPS2013_5071}.

Knowledge graphs are collections of real-world triples,  where each triple or fact $(h, r, t)$ in KGs represents some relation  $r$ between a head  entity $h$ and a tail  entity $t$. KGs can thus be formalized as directed multi-relational graphs, where nodes correspond
to entities and edges linking the nodes encode various kinds of relationships \citep{Garcia-DuranBUG15,NickelMTG15}.  Here entities are real-world things or objects such as  persons, places, organizations, music tracks or movies. Each relation type defines a certain relationship between entities. For example, as illustrated in Figure \ref{fig:kbexample}, the relation type ``$\mathsf{child\_of}$'' relates person entities with each other, while the  relation type ``$\mathsf{born\_in}$'' relates person entities with place  entities.  
Several KG examples  include the domain-specific KG GeneOntology  and popular generic KGs of WordNet \citep{FellbaumC98},  YAGO
\citep{Suchanek:2007}, Freebase \citep{Bollacker:2008}, NELL \citep{Carlson:2010:TAN:2898607.2898816} and DBpedia
\citep{LehmannIJJKMHMK15} as well as commercial KGs such as Google's Knowledge Graph, Microsoft's Satori and Facebook's Open Graph.  Nowadays, KGs  are used in a  number of commercial applications including  search engines such as Google, Microsoft's Bing and Facebook's Graph search. They also  are useful resources for many natural language processing tasks such as question answering  \citep{Ferrucci:2012:ITW:2481742.2481743,Fader:2014:OQA:2623330.2623677}, word sense disambiguation \citep{Navigli:2005:SSI:1070615.1070793,Agirre2013}, semantic parsing \citep{krishnamurthy-mitchell:2012:EMNLP-CoNLL,berant-EtAl:2013:EMNLP} and co-reference resolution \citep{ponzetto-strube:2006:HLT-NAACL06-Main,TACL522}.

A main issue is that even very large KGs, such as Freebase and DBpedia, which contain billions of fact triples about the world,  are still far from complete. In particular, in English DBpedia 2014, 60\% of person entities miss a place of birth and 58\% of the scientists do not have a fact about what they are known for \citep{Denis2015}.  In Freebase, 71\% of 3 million person entities miss a place of birth, 75\% do not have a nationality while 94\% have no facts about their parents  \citep{West:2014:KBC:2566486.2568032}. So, in terms of a specific application,  question answering systems based on  incomplete KGs would not provide a correct answer given a correctly interpreted question. For example, given the incomplete KG in Figure \ref{fig:kbexample}, it would be impossible to answer the question ``where was Jane born ?'', although the question is completely matched with existing entity and relation type information (i.e. ``$\mathsf{Jane}$'' and ``$\mathsf{born\_in}$'') in KG. 
 Consequently, much work has  been devoted towards  knowledge graph completion to perform link prediction in KGs, which attempts to predict whether a relationship/triple not in the KG is likely to be true, i.e. to add new triples by leveraging existing triples in the KG  \citep{Lao2010,Bordes2014SME,Gardner2014,Garcia-DuranBUG15}.  For example, we would like to predict the missing tail entity in the incomplete triple $\mathsf{(Jane, born\_in, ?)}$ or  predict whether the triple $\mathsf{(Jane, born\_in, Miami)}$ is correct or not.

Embedding models for KG completion have been proven to give state-of-the-art link prediction performances, in which entities are represented by latent feature vectors while relation types are represented by latent feature vectors and/or matrices and/or third-order tensors  
\citep{NIPS2013_5071,NIPS2013_5028}.  This paper: (1) surveys the embedding models for KG completion, then (2) summarizes up-to-date experimental results on the  standard   evaluation task of  entity prediction---which is also referred to as the {link prediction} task \citep{NIPS2013_5071}, and (3)  points out potential future research directions.

\section{A General Approach of Embedding Models for KG Completion}

 Let $\mathcal{E}$ denote the set of entities and $\mathcal{R}$ the set
of relation types. 
Denote by $\mathcal{G}$ the knowledge graph consisting of a set of correct triples $(h, r, t)$, such that $h, t \in \mathcal{E}$ and $r \in \mathcal{R}$.
For each triple $(h, r, t)$, the embedding models define a \emph{score
function} $f(h, r, t)$ of its plausibility.
Their goal here is to:

\begin{description}
\item \textbf{Choose $f$ such that the score $f(h, r, t)$ of a correct  triple $(h, r, t)$ is higher than the score $f(h', r', t')$ of an incorrect triple $(h',r',t')$.} 
\end{description}

For example, TransE defines a score function of $f_{\text{TransE}}(h, r, t) = - \|  \boldsymbol{v}_{h} + \boldsymbol{v}_{r} - \boldsymbol{v}_{t} \|$, where $h$, $r$ and $t$ are represented by low dimensional  vectors $\boldsymbol{v}_{h}$, $\boldsymbol{v}_{r}$ and $\boldsymbol{v}_{t}$, respectively. As  $\mathsf{(Tokyo, is\_capital\_of, Japan)}$ is a correct triple, while $\mathsf{(Tokyo, is\_capital\_of, Portugal)}$ and $\mathsf{(Lisbon, is\_capital\_of, Japan)}$ are incorrect ones, we would have: $-\|\boldsymbol{v}_{Tokyo} + \boldsymbol{v}_{{is\_capital\_of}} - \boldsymbol{v}_{Japan}\| > -\|\boldsymbol{v}_{Tokyo} + \boldsymbol{v}_{{is\_capital\_of}} - \boldsymbol{v}_{Portugal}\|$,  and $
-\|\boldsymbol{v}_{Tokyo} + \boldsymbol{v}_{{is\_capital\_of}} - \boldsymbol{v}_{Japan}\| > -\|\boldsymbol{v}_{Lisbon} + \boldsymbol{v}_{{is\_capital\_of}} - \boldsymbol{v}_{Japan}\|$. Table \ref{tab:emmethods} in Section \ref{sec:models}  summarizes different prominent score
functions $f(h, r, t)$. 


 To learn model parameters (i.e. entity vectors, relation vectors or matrices), the embedding models  minimize an objective loss $\mathcal{L}$. A conventional objective loss is the margin-based  pairwise ranking loss \citep{NIPS2013_5071}:

\begin{align*}
\mathcal{L}_{\text{Margin}} =  \sum_{\substack{(h,r,t) \in \mathcal{G} \\ (h',r,t') \in \mathcal{G}'_{(h, r, t)}}} [\gamma - f(h, r, t) + f(h', r, t')]_+  
 \label{equal:objfunc}
  \end{align*}

\noindent where $[x]_+ = \max(0, x)$; $\gamma$ is the margin hyper-parameter; and $\mathcal{G}'_{(h, r, t)} $ is the set of incorrect triples generated by corrupting the correct triple $(h, r, t)\in\mathcal{G}$. 

Also, the negative log-likelihood (NLL) of  softmax regression   \citep{toutanova-chen:2015:CVSC}  and the NLL of logistic regression   \citep{TrouillonWRGB16} are commonly used in recent KG completion  research:\footnote{All the losses can also include an L2 regularization on the model parameters, which is not shown for simplification.}  

\begin{eqnarray*}
 \mathcal{L}_{\text{Softmax}} &=& -\sum_{\substack{(h, r, t) \in \mathcal{G}}} \Bigg(\dfrac{\exp\big(f\left(h, r, t\right)\big)}{\sum\limits_{\substack{t'\ \in\ \mathcal{E} \setminus \{t\}}} \exp\big(f\left(h, r, t'\right)\big) } \\ \nonumber 
 && \ \ \ \ \ \ \ \ \ \ \ \ \ \ \ \ \ \ \ \ \ + \dfrac{\exp\big(f\left(h, r, t\right)\big)}{\sum\limits_{\substack{h'\ \in\ \mathcal{E} \setminus \{h\}}} \exp\big(f\left(h', r, t\right)\big) } \Bigg)  \\ 
\nonumber \\
 \mathcal{L}_{\text{Logistic}} &=&  \sum_{\substack{(h, r, t) \in \{\mathcal{G} \cup \mathcal{G}'\}}} \log\left(1 + \exp\left(- \text{I}_{(h, r, t)} \cdot f\left(h, r, t\right)\right)\right)  \\
 & & \text{with: } \text{I}_{(h, r, t)} = \left\{ 
  \begin{array}{l}
  1\;\text{for } (h, r, t)\in\mathcal{G}\\
 -1\;\text{for } (h, r, t)\in\mathcal{G}'
  \end{array} \right.  \nonumber
\end{eqnarray*}

To corrupt the head or tail entities, a common strategy is to  uniformly replace the entities  when sampling  incorrect triples \citep{NIPS2013_5071}, however it  results in many false negative labels \cite{AAAI148531}. Domain sampling \cite{Denis2015,xie-EtAl:2017:Long} generates corrupted triples by sampling entities from the same domain or from the set of relation-dependent entities. 
The ``{Bernoulli}'' trick \cite{AAAI148531} is widely used to set  different probabilities 
for generating head or tail entities: 
For each relation type $r$,  we calculate the
averaged number $a_{r,1}$ of heads $h$ for a pair $(r, t)$ and the averaged
number $a_{r,2}$ of tails $t$ for a pair $(h, r)$. We then define a Bernoulli distribution  with success probability $\lambda_r = \dfrac{a_{r,1}}{a_{r,1} + a_{r,2}}$ for sampling: given a correct triple $(h, r, t)$, we corrupt this triple by replacing head entity with probability  $\lambda_r$ while replacing the tail entity with probability  $(1 - \lambda_r)$. 

Recently, \newcite{Cai2017} and \newcite{sun2018rotate} proposed adversarial learning-based strategies for sampling incorrect triples. However, they did not provide a comparison between the adversarial learning-based strategies and the ``{Bernoulli}'' trick.

%

\section{Specific Models}\label{sec:models}

\subsection{Triple-based Embedding Models}\label{ssec:triplemdels}

\paragraph{Translation-based models:} The {Unstructured} model \citep{Bordes2014SME} assumes that the head and tail entity vectors are similar. As the Unstructured model does not take the relationship into account, it cannot distinguish different relation types. The {Structured Embedding} (SE) model \citep{bordes-2011} assumes that the head and tail entities are similar only in a relation-dependent subspace, where each relation is represented by two different matrices. 
 {TransE} 
\citep{NIPS2013_5071} is inspired by models such as the Word2Vec Skip-gram model 
\citep{MikolovNIPS2013} where relationships between words often correspond
to translations in latent feature space.  In particular, 
TransE learns low-dimensional and dense vectors for every entity and relation type, so that 
each relation type corresponds to a translation vector operating on the vectors representing the entities, i.e. $\boldsymbol{v}_{h} + \boldsymbol{v}_{r} \approx \boldsymbol{v}_{t}$ for each fact triple $(h, r, t)$.  
TransE thus is suitable for 1-to-1 relationships, such as ``$\mathsf{is\_capital\_of}$'', where a head entity is linked to at most one tail entity given a relation type. Because of using only one translation vector to represent  each relation type, TransE is not well-suited for Many-to-1, 1-to-Many and Many-to-Many relationships,\footnote{A relation type $r$ is classified    Many-to-1 if  multiple head entities can be connected by $r$ to at most one tail entity. A relation type $r$ is classified    1-to-Many if  multiple tail entities can be linked by $r$ from at most one head entity. A relation type $r$ is classified    Many-to-Many  if  multiple head entities can be connected by $r$ to a tail entity and vice versa.} such as for relation types ``$\mathsf{born\_in}$'', ``$\mathsf{place\_of\_birth}$'' and  ``$\mathsf{research\_fields}$.'' For example in Figure \ref{fig:kbexample},   using one vector representing the  relation type ``$\mathsf{born\_in}$'' cannot capture both the translating direction from ``$\mathsf{Patti}$'' to ``$\mathsf{Miami}$'' and its inverse direction from ``$\mathsf{Mom}$'' to ``$\mathsf{Austin}$.''

 \begin{table*}[!t]
\centering
\resizebox{16cm}{!}{
\def\arraystretch{1.33}
\setlength{\tabcolsep}{0.4em}
\begin{tabular}{l|l|l}
\hline
\multicolumn{2}{c|}{\textbf{Model}} & \bf Score function $f(h, r, t)$  \\
\hline
\multirow{9}{*}{\rotatebox[origin=c]{90}{Translation}} & Unstructured &  $- \| \boldsymbol{v}_{h} - \boldsymbol{v}_{t} \|_{\ell_{1/2}}$  \\
\cline{2-3}
& SE &  $-\| \textbf{W}_{r,1}\boldsymbol{v}_{h} - \textbf{W}_{r,2}\boldsymbol{v}_{t}\|_{\ell_{1/2}}$\ \ where     $\textbf{W}_{r,1}$, $\textbf{W}_{r,2}$ $\in$ $\mathbb{R}^{k \times k}$  \\
\cline{2-3}
& TransE &  $- \|  \boldsymbol{v}_{h} + \boldsymbol{v}_{r} - \boldsymbol{v}_{t} \|_{\ell_{1/2}}$\ \ where   $\boldsymbol{v}_{r}  \in  \mathbb{R}^{k}$  \\
\cline{2-3}
& {TransH} & $-\| (\textbf{I} - \boldsymbol{r}_{p}\boldsymbol{r}_{p}^{\top})\boldsymbol{v}_{h} + \boldsymbol{v}_{r} - (\textbf{I} - \boldsymbol{r}_{p}\boldsymbol{r}_{p}^{\top})\boldsymbol{v}_{t} \|_{\ell_{1/2}}$\ \ where   $\boldsymbol{r}_{p}$, $\boldsymbol{v}_{r} \in$ $\mathbb{R}^{k}$ , $\textbf{I}$ denotes an identity matrix size $k \times k$  \\
\cline{2-3}
& {TransR} & $-\| \textbf{W}_{r}\boldsymbol{v}_{h} + \boldsymbol{v}_{r} - \textbf{W}_{r}\boldsymbol{v}_{t}\|_{\ell_{1/2}}$\ \ where  $\textbf{W}_{r}$ $\in$ $\mathbb{R}^{n \times k}$ ,    $\boldsymbol{v}_{r}$ $\in$ $\mathbb{R}^{n}$  \\
\cline{2-3}
& {STransE} & $-\| \textbf{W}_{r,1}\boldsymbol{v}_{h} + \boldsymbol{v}_{r} - \textbf{W}_{r,2}\boldsymbol{v}_{t}\|_{\ell_{1/2}}$\ \ where  $\textbf{W}_{r,1}$, $\textbf{W}_{r,2}$ $\in$ $\mathbb{R}^{k \times k}$ , $\boldsymbol{v}_{r} \in \mathbb{R}^{k}$  \\
\cline{2-3}
& {TranSparse} & $-\| \textbf{W}_{r,1}(\theta_{r,1})\boldsymbol{v}_{h}+ \boldsymbol{v}_{r} - \textbf{W}_{r,2}(\theta_{r,2})\boldsymbol{v}_{t}\|_{\ell_{1/2}}$\ \ where   $\textbf{W}_{r,1}$, $\textbf{W}_{r,2}$ $\in$ $\mathbb{R}^{n \times k}$\ \ ;\ \   $\theta_{r,1}$, $\theta_{r,2} \in \mathbb{R}$\ \ ;\ \   $\boldsymbol{v}_{r}$ $\in$ $\mathbb{R}^{n}$  \\
\cline{2-3}
& {TransD} & $-\| (\textbf{I} + \boldsymbol{r}_{p}\boldsymbol{h}_{p}^{\top})\boldsymbol{v}_{h} + \boldsymbol{v}_{r} - (\textbf{I} + \boldsymbol{r}_{p}\boldsymbol{t}_{p}^{\top})\boldsymbol{v}_{t} \|_{\ell_{1/2}}$\ \ where  $\boldsymbol{r}_{p}$, $\boldsymbol{v}_{r}$,  $\boldsymbol{h}_{p}, \boldsymbol{t}_{p}$ $\in$ $\mathbb{R}^{k}$  \\
\cline{2-3}
& {lppTransD} & $-\| (\textbf{I} + \boldsymbol{r}_{p,1}\boldsymbol{h}_{p}^{\top})\boldsymbol{v}_{h} + \boldsymbol{v}_{r} - (\textbf{I} + \boldsymbol{r}_{p,2}\boldsymbol{t}_{p}^{\top})\boldsymbol{v}_{t} \|_{\ell_{1/2}}$\ \ where  $\boldsymbol{r}_{p,1}$, $\boldsymbol{r}_{p,2}$, $\boldsymbol{v}_{r}$, $\boldsymbol{h}_{p}, \boldsymbol{t}_{p}$ $\in$ $\mathbb{R}^{k}$   \\
\hline
\multirow{6}{*}{\rotatebox[origin=c]{90}{Bilinear \& Tensor}} & Bilinear &  $ \boldsymbol{v}_{h}^{\top}\textbf{W}_{r}\boldsymbol{v}_{t}$\ \ where     $\textbf{W}_{r}$  $\in$ $\mathbb{R}^{k \times k}$  \\
\cline{2-3}
& DISTMULT &  $ \boldsymbol{v}_{h}^{\top}\textbf{W}_{r}\boldsymbol{v}_{t}$\ \ where    $\textbf{W}_{r}$ is a diagonal matrix $\in$ $\mathbb{R}^{k \times k}$ \\
\cline{2-3}
& {SimplE} & $\frac{1}{2}$\big($ \boldsymbol{v}_{h,1}^{\top}\textbf{W}_{r}\boldsymbol{v}_{t,2}$ + $ \boldsymbol{v}_{t,1}^{\top}\textbf{W}_{r^{-1}}\boldsymbol{v}_{h,2}$\big)\ \ where  $\boldsymbol{v}_{h,1}, \boldsymbol{v}_{h,2}, \boldsymbol{v}_{t,1}, \boldsymbol{v}_{t,2} \in \mathbb{R}^k$ \ ; \ $\textbf{W}_{r}$ and $\textbf{W}_{r^{-1}}$   are diagonal matrices $\in$ $\mathbb{R}^{k \times k}$  \\
\cline{2-3}
& {SME(bilinear)} & $\boldsymbol{v}_{h}^{\top}(\textbf{M}_{1}\times_3\boldsymbol{v}_{r})^{\top}(\textbf{M}_{2}\times_3\boldsymbol{v}_{r})\boldsymbol{v}_{t}$\ \ where  $\boldsymbol{v}_r \in \mathbb{R}^k$\ \ ;\ \   $\textbf{M}_{1}, \textbf{M}_{2} \in \mathbb{R}^{n \times k \times k}$\\


 
\cline{2-3}
& TuckER & $\textbf{M}\times_1\boldsymbol{v}_{h}\times_2\boldsymbol{v}_{r}\times_3\boldsymbol{v}_{t}$\ \ where  $\boldsymbol{v}_r \in \mathbb{R}^n$\ ,\   $\textbf{M} \in \mathbb{R}^{k \times n \times k}$ \ ; \ $\times_d$ denotes the tensor product along the $d$-th mode\\

\cline{2-3}
& HolE & $\mathsf{sigmoid}(\boldsymbol{v}^{\top}_t(\boldsymbol{v}_h \star  \boldsymbol{v}_r))$\ \ where  $\star\ \text{denotes circular correlation}$ \\ 

\hline
\multirow{4}{*}{\rotatebox[origin=c]{90}{Neural network}}& {NTN} & $ \boldsymbol{v}_r^{\top} \mathsf{tanh}( \boldsymbol{v}_{h}^{\top}\textbf{M}_{r}\boldsymbol{v}_{t}  + \textbf{W}_{r,1}\boldsymbol{v}_{h} + \textbf{W}_{r,2}\boldsymbol{v}_{t} + \textbf{b}_r)$\ \ where   $\boldsymbol{v}_r\text{, } \textbf{b}_r \in \mathbb{R}^n$\ \ ;\ \   $\textbf{M}_{r} \in \mathbb{R}^{k \times k \times n}$\ \ ;\ \    $\textbf{W}_{r,1}$, $\textbf{W}_{r,2} \in \mathbb{R}^{n \times k}$ \\

\cline{2-3}
& ER-MLP & $\mathsf{sigmoid}(\bold{w}^{\top}\mathsf{tanh}(\boldsymbol{W}\mathsf{concat}(\boldsymbol{v}_h , \boldsymbol{v}_r , \boldsymbol{v}_t)))$\\

\cline{2-3}

& ConvE & $\boldsymbol{v}^{\top}_t \mathsf{ReLU}\left(\boldsymbol{W}\mathsf{vec}\left(\mathsf{ReLU}\left(\mathsf{concat} (\overline{\boldsymbol{v}}_h , \overline{\boldsymbol{v}}_r)\ast\bold{\Omega}\right)\right)\right)$
   \\
\cline{2-3}
& ConvKB & $\bold{w}^{\top} \mathsf{concat}\left(\mathsf{ReLU}\left([\boldsymbol{v}_h , \boldsymbol{v}_r , \boldsymbol{v}_t]\ast\bold{\Omega}\right)\right)$  \\

\hline
\multirow{4}{*}{\rotatebox[origin=c]{90}{Complex vector}} & {ComplEx} & $\mathsf{Re}\left(\boldsymbol{c}_{h}^{\top}\textbf{C}_{r}\hat{\boldsymbol{c}}_{t}\right)$\ \ where  $\mathsf{Re}(c)$ denotes the real part of the complex value $c \in \mathbb{C}$  \\
&  & $\boldsymbol{c}_{h}, {\boldsymbol{c}}_{t} \in \mathbb{C}^{k}$\ \ ;\ \   $\textbf{C}_{r} \in \mathbb{C}^{k\times k}$ is a diagonal matrix \ ; \  $\hat{\boldsymbol{c}}_t$ is the conjugate of $\boldsymbol{c}_t$ \\
\cline{2-3}
& {RotatE} &   $-\| \boldsymbol{c}_{h} \circ \boldsymbol{c}_{r} - \boldsymbol{c}_{t}\|_{\ell_{1/2}}$\ \ where $\boldsymbol{c}_{h}, \boldsymbol{c}_{r}, \boldsymbol{c}_{t} \in  \mathbb{C}^{k}$\ \ ;\ \   $\circ$ denotes the element-wise product  \\
\cline{2-3} 
& QuatE & $\boldsymbol{q}_{h} \otimes \dfrac{\boldsymbol{q}_{r}}{|\boldsymbol{q}_{r}|} \bullet \boldsymbol{q}_{t} $\ \ where $\boldsymbol{q}_{h}, \boldsymbol{q}_{r}, \boldsymbol{q}_{t} \in  \mathbb{H}^{k}$\ \ ;\ \   $\otimes$ and $\bullet$ denote Hamilton and quaternion inner products, respectively\\

\hline

\multirow{2}{*}{\rotatebox[origin=c]{90}{Path}}& TransE-\textsc{comp} &  $ -\|  \boldsymbol{v}_{h} + \boldsymbol{v}_{r_1} + \boldsymbol{v}_{r_2} + ... + \boldsymbol{v}_{r_m} - \boldsymbol{v}_{t} \|_{\ell_{1/2}}$\ \ where  $\boldsymbol{v}_{r_1}, \boldsymbol{v}_{r_2} ,..., \boldsymbol{v}_{r_m} \in \mathbb{R}^{k}$  \\
\cline{2-3}
& Bilinear-\textsc{comp} &  $ \boldsymbol{v}_{h}^{\top}\textbf{W}_{r_1}\textbf{W}_{r_2}...\textbf{W}_{r_m}\boldsymbol{v}_{t}$\ \ where    $\textbf{W}_{r_1}, \textbf{W}_{r_2},..., \textbf{W}_{r_m} \in \mathbb{R}^{k \times k}$  \\
\hline
\end{tabular}
}
\caption{The score functions $f(h,r, t)$ of several prominent embedding models for KG completion.  
  In these models, the entities $h$ and $t$ are represented
  by  vectors $\boldsymbol{v}_{h}$ and $\boldsymbol{v}_{t} \in \mathbb{R}^{k}$, respectively. $\ell_{1/2}$ denotes either the L$_1$-norm or the squared L$_2$-norm. 
  In ConvE, $\overline{\boldsymbol{v}}_h$ and $\overline{\boldsymbol{v}}_r$  denote a 2D reshaping of  $\boldsymbol{v}_h$ and  $\boldsymbol{v}_r$, respectively. In both ConvE and ConvKB models, $\ast$ and $\bold{\Omega}$ denote a convolution operator and  a set of filters, respectively.
}
\label{tab:emmethods}
\end{table*}

To overcome those issues of TransE,  TransH \citep{AAAI148531} associates each relation with a
relation-specific hyperplane and uses a projection vector to project
entity vectors onto that hyperplane. TransD
\citep{ji-EtAl:2015:ACL-IJCNLP} and TransR/CTransR \citep{AAAI159571}
extend  TransH  by using two projection vectors and a matrix to project entity vectors into a relation-specific space, respectively. Similar to TransR, TransR-FT \citep{FengHWZHZ16} also uses  a matrix to project head and tail entity vectors. 
TEKE\_H  \citep{DBLP:conf/ijcai/WangL16}  extends TransH to  incorporate rich context information in an external text corpus. 
lppTransD \citep{yoon-EtAl:2016:N16-1} extends TransD to additionally use two projection vectors for representing each relation.   STransE  \citep{NguyenNAACL2016} and  TranSparse   \citep{JiLH016} can be  viewed as direct extensions of  TransR, where head and tail entities are associated with their own projection matrices. Unlike  STransE,  TranSparse uses adaptive sparse matrices, whose sparse degrees are defined based on the number of entities linked by  relations. TranSparse-DT \citep{8057770} is an extension of TranSparse with a dynamic translation. ITransF \citep{xie-EtAl:2017:Long}  can be considered as a generalization of STransE, which allows the sharing of statistic regularities between relation projection matrices  and alleviates data sparsity issue. Furthermore, TorusE \citep{Ebisu2018}  embeds entities and relations on a torus to handle TransE's regularization problem which  forces entity embeddings to be on a sphere in the embedding vector space. 

\paragraph{Bilinear- \& Tensor-based models:} DISTMULT  \citep{yang-etal-2015} is based on the {Bilinear} model \citep{ICML2011Nickel_438,NIPS2012_4744} where each relation is represented by a diagonal matrix rather than a full matrix. SimplE \citep{NIPS2018_7682} extends DISTMULT  to allow
 two embeddings of each entity to be learned dependently. Such quadratic   forms are also used to model entities and relations in KG2E \citep{He:2015},   TATEC \citep{Garcia-DuranBUG15}, TransG \citep{xiao-huang-zhu:2016:P16-1}, RSTE \citep{Tay:2017:RST:3018661.3018695},  ANALOGY \citep{pmlr-v70-liu17d} and Dihedral \cite{xu-li-2019-relation}. SME-bilinear  \citep{Bordes2014SME} is proposed to first separately combine entity-relation pairs $(h,r)$ and $(r,t)$ and then semantically match these combinations, using tensor product.  HolE   \citep{Nickel:2016:HEK:3016100.3016172} uses  circular correlation--a compositional operator--which can be interpreted as a compression of the tensor product. In addition,  
TuckER \citep{balazevic-etal-2019-tucker} is a linear model based on the Tucker  tensor  decomposition of the binary tensor representation of KG  triples.

\paragraph{Neural network-based models:} The neural tensor network (NTN) model \citep{NIPS2013_5028} also uses a bilinear tensor operator to represent each relation while  ProjE \citep{ShiW16a}  can be viewed as simplified versions of NTN. The 
ER-MLP model \citep{Dong:2014} represents each triple by a vector obtained from concatenating head, relation and tail embeddings, then feeds this vector into a single-layer MLP with one-node output layer. 
ConvE \citep{DettmersMSR17} and ConvKB \citep{NguyenNNP2017} are based on convolutional neural networks. ConvE uses a  convolution layer directly over 2D reshaping of head-entity and relation embeddings, while ConvKB applies a convolution layer over the embedding triples (here each triple $(h,r,t)$ is represented as a 3-column matrix where each column vector represents a triple element).  HypER \cite{balazevic2019hypernetwork} simplifies ConvE by using  a hypernetwork to produce 1D convolutional filters for each relation, then extracts relation-specific features from head entity embeddings.  Conv-TransE \citep{Shang19} extends ConvE to keep the translational characteristic between entities and relations.   InteractE  \citep{interacte2020} uses a circular convolution operator and a checkered reshaping function instead of the standard convolution operator and 2D stack reshaping function in ConvE.  
The CapsE model \citep{NguyenVNNP_NAACL2019} extends ConvKB by stacking a capsule network
layer \citep{NIPS2017_6975} on top of the convolution layer.

\paragraph{Complex vector-based models:} Instead of embedding entities and relations in the real-valued vector space,  ComplEx \citep{TrouillonWRGB16} is an extension of DISTMULT in the complex vector space.  ComplEx-N3 \citep{LacroixUO18}  extends ComplEx with
weighted nuclear 3-norm. 
 Also in the complex vector space, RotatE \citep{sun2018rotate}
 defines each relation as a rotation from the head entity to the tail entity. QuatE \citep{NIPS2019_8541} represents entities by quaternion embeddings (i.e. hypercomplex-valued embeddings) and models relations as rotations in the quaternion space by employing the Hamilton and quaternion-inner  products.

\subsection{Relation Path-based Embedding Models}

All embedding models mentioned above in Section \ref{ssec:triplemdels} only take triples into account. Thus, these models ignore potentially useful information implicitly presented by the structure of the KG. For example, the relation path $h \xrightarrow{\mathsf{born\_in\_city}} e \xrightarrow{\mathsf{city\_in\_country}} t$ should indicate a  relationship ``$\mathsf{nationality}$'' between the $h$ and $t$ entities. Also, neighborhood information of entities  could be useful for predicting the relationship between two entities as well. For example, in the KG NELL \citep{Carlson:2010:TAN:2898607.2898816}, we have information such as if a person works for  an organization and  this person also leads that organization, then it is likely that this person is the CEO of that organization. 

Recent research has also shown that relation paths between entities in KGs provide richer context information and improve the performance of embedding models for KG completion \citep{luo-EtAl:2015:EMNLP3,LiangF15,garciaduran-bordes-usunier:2015:EMNLP,guu-miller-liang:2015:EMNLP,toutanova-EtAl:2016:P16-1,Alberto17,P18-1200,N18-1165}. In particular, 
\newcite{luo-EtAl:2015:EMNLP3} constructed relation paths between entities and, viewing entities and relations in the path as pseudo-words,  then applied  Word2Vec \citep{MikolovNIPS2013} to produce pre-trained vectors for these pseudo-words. 
\newcite{luo-EtAl:2015:EMNLP3} showed that using these pre-trained vectors for initialization helps to improve the performance of  models TransE \citep{NIPS2013_5071},   SME  \citep{Bordes2014SME} and SE  \citep{bordes-2011}.    \newcite{LiangF15}   used the plausibility score produced by SME to compute the weights of relation paths.  

PTransE-RNN \citep{lin-EtAl:2015:EMNLP1} models  relation paths by using a recurrent neural network (RNN). In addition, \newcite{das-EtAl:2017:EACLlong1}'s model  and ROPs \citep{yin-etal-2018-recurrent} also apply RNN to model the path between an entity pair, however,  in contrast to PTransE-RNN, they  additionally take the intermediate entities present in the path into account. 
IRN   \citep{shen-EtAl:2017:RepL4NLP1} uses a  shared memory and RNN-based controller to implicitly model multi-step structured relationships. 
\textsc{r}TransE \citep{garciaduran-bordes-usunier:2015:EMNLP}, PTransE-ADD \citep{lin-EtAl:2015:EMNLP1} and TransE-\textsc{comp} \citep{guu-miller-liang:2015:EMNLP} extend  TransE  to represent a relation path by a vector which is the sum of the vectors of all relations in the path. In  Bilinear-\textsc{comp}  \citep{guu-miller-liang:2015:EMNLP} and  \textsc{pruned}-\textsc{paths}  \citep{toutanova-EtAl:2016:P16-1}, each relation is a matrix and so it represents the relation path by matrix multiplication.  \newcite{Alberto17} proposed the KB$_{LRN}$  framework to combine relational paths with latent and numerical features.

The neighborhood mixture model TransE-NMM  \citep{NguyenCoNLL2016} can be also viewed as a three-relation path model because it takes into account the neighborhood entity and relation information of both head and tail entities in each triple. 
 ReInceptionE \cite{xie-etal-2020-reinceptione} employs the Inception network \cite{7780677} to increase the interactions between head and relation embeddings for obtaining better representations of the head and relation pairs and then uses a relation-aware attention mechanism to enrich these pair representations with the local neighborhood and global entity information. 
Neighborhood information is also exploited in R-GCN \citep{SchlichtkrullKB17}, SACN \citep{Shang19} and KBGAT \citep{nathani-etal-2019-learning}, which generalize graph convolutional networks \citep{KipfW17} and graph attention networks \citep{velickovic2018graph} for dealing with highly multi-relational data, e.g.  KGs. 
For computing the final representation of an entity, they make use of  layer-wise propagation to accumulate linearly-transformed embeddings of its neighboring entities  through a normalized sum with different relational  weights. For link prediction, R-GCN,  SACN and KBGAT apply DISTMULT, Conv-TransE and ConvKB to compute triple scores, respectively.

\subsection{Other KG Completion Models} 
The Path Ranking Algorithm (PRA) \citep{Lao2010} is a random walk inference technique which was proposed to predict a new relationship  between two entities in KGs. 
\newcite{Lao:2011:RWI:2145432.2145494} used PRA to estimate  the probability of an unseen triple as a combination of  weighted  random walks that follow different paths  linking the head entity and tail entity  in the KG. \newcite{Gardner2014} made use of an external text corpus to  increase the connectivity of the KG used as the input to PRA. \newcite{gardner-mitchell:2015:EMNLP} improved PRA by proposing a subgraph feature extraction technique to make the generation of random walks in KGs more efficient and expressive, while \newcite{wang-EtAl:2016:P16-13} extended PRA to couple the path ranking of multiple relations. PRA can also be used  in conjunction with first-order logic in the discriminative Gaifman model \citep{NIPS2016_6098}. 
In addition, \newcite{neelakantan-roth-mccallum:2015:ACL-IJCNLP} used a RNN to learn vector representations of PRA-style relation paths between entities in the KG. Other random-walk based learning algorithms for  KG completion can be also found in \newcite{feng-EtAl:2016:COLING1}, \newcite{Liu:2016:HRW:2911451.2911509}, \newcite{wei-zhao-liu:2016:EMNLP2016}, \newcite{ijcai2017-166} and \newcite{das2018go}. 

\newcite{NIPS2017_6826} proposed a Neural Logic Programming (LP) framework to learning probabilistic first-order logical rules for KG reasoning, producing competitive link prediction performances.  \newcite{commonsense2019} presented an approach to generate sentences from triples via hand-craft templates, and  then use the likelihoods produced by the pre-trained BERT \citep{devlin-etal-2019-bert}  for these generated sentences to score the plausibility of the corresponding triples. 
 See  other methods for learning from KGs and multi-relational data in \newcite{NickelMTG15} and \newcite{8047276}.

\section{Evaluation Task}

The standard evaluation task of entity prediction, i.e. the link prediction task  \citep{NIPS2013_5071},  is proposed to evaluate embedding models for KG completion.\footnote{Another evaluation task for KG completion is triple classification  \citep{NIPS2013_5028}, however, it is not as widely used as the link prediction task. See the Appendix  for a summary of state-of-the-art triple classification results.}

\paragraph{Datasets:}\ 
Information about benchmark datasets for KG completion evaluation is given in Table \ref{tab:datasets}. 
FB15k and WN18 are derived from the large real-world KG Freebase \cite{Bollacker:2008} and  the large lexical KG WordNet \cite{Miller:1995:WLD:219717.219748}, respectively.  
\newcite{toutanova-chen:2015:CVSC}  noted that FB15k and WN18 are not challenging datasets because they contain many reversible triples. \newcite{DettmersMSR17}  showed a concrete example: A test triple ($\mathsf{feline, hyponym, cat}$) can be mapped to a training triple ($\mathsf{cat, hypernym, feline}$), thus knowing that ``$\mathsf{hyponym}$'' and ``$\mathsf{hypernym}$'' are reversible  allows us to easily predict the majority of test triples. So, datasets  FB15k-237 \cite{toutanova-chen:2015:CVSC}  and WN18RR \cite{DettmersMSR17} are created to serve as realistic KG completion datasets which represent a more challenging learning setting.  FB15k-237 and WN18RR are subsets of  FB15k and WN18, respectively. 

\subsection{Task Description}

The entity prediction task, i.e. link prediction
\cite{NIPS2013_5071},  predicts the head or the tail entity given the relation type and
the other entity, i.e. predicting $h$ given $(?, r, t)$ or predicting
$t$ given $(h, r, ?)$ where $?$ denotes the missing element.
The results are evaluated using a ranking  induced by the function $f(h, r, t)$ on test triples. 

Each correct test triple $(h, r, t)$ is corrupted by replacing either its head or tail entity by each of the possible 
entities in turn, and then these candidates are ranked in descending order of
their plausibility score.  
The ``Filtered'' setting protocol, described in \newcite{NIPS2013_5071}, 
 filters out before ranking any corrupted triples
that appear in the KG. Ranking a corrupted triple appearing in the KG  (i.e. a correct triple)  higher than the original test triple is also correct, thus this ``Filtered'' setting provides a clear view on the ranking performance.

In addition to the mean rank and  the Hits@10 (i.e. the proportion of
test triples for which the target entity is ranked in
the top 10 predictions), which were originally used in the entity prediction task \cite{NIPS2013_5071}, recent work also reports the mean reciprocal rank  (\textbf{MRR}).\footnote{See  \newcite{Ricardo2011} for  definitions of the mean rank, Hits@10 and MRR. Some recent work additionally reported Hits@1 (i.e.  the  proportion  of  test  triples  for  which  the target entity is ranked first). However, formulas of
MRR and Hits@1 show a strong correlation between these two scores. So using Hits@1 might not reveal any additional insight. } Mean rank is always greater or equal to 1  and the
lower mean
rank indicates better entity prediction performance, while MRR and Hits@10 scores always range from 0.0 to 1.0, and higher score reflects better prediction result. 

\begin{table}[!t]
\centering
\resizebox{12cm}{!}{
\begin{tabular}{l|lllll}
\hline
\bf Dataset & $\mid\mathcal{E}\mid$ & $\mid\mathcal{R}\mid$  & \multicolumn{3}{l}{\#Triples in train/valid/test} \\
\hline
FB15k \cite{NIPS2013_5071} & 14,951 & 1,345 & 483,142 & 50,000 & 59,071\\
WN18 \cite{NIPS2013_5071} & 40,943 & 18 & 141,442 & 5,000 & 5,000\\
FB15k-237 \cite{toutanova-chen:2015:CVSC} & 14,541 & 237 & 272,115 & 17,535 & 20,466 \\
WN18RR  \newcite{DettmersMSR17} & 40,943 & 11 & 86,835 & 3,034 & 3,134 \\
\hline
\end{tabular}
}
\caption{Statistics of  benchmark experimental datasets. }
\label{tab:datasets}
\end{table}

\begin{table}[!t]
\centering
\resizebox{10cm}{!}{
\begin{tabular}{l|lll|lll}
\hline
\multirow{3}{*}{\bf Method}& \multicolumn{6}{|c}{\textbf{Filtered}} \\
\cline{2-7}
& \multicolumn{3}{|c}{\bf FB15k} & \multicolumn{3}{|c}{\bf WN18}\\
\cline{2-7}
 &   MR & @10 & MRR&   MR & @10 & MRR \\
 \hline
TransH \citep{AAAI148531}  &  87 &  64.4&-&  303 & 86.7 & - \\
TransR \citep{AAAI159571}  &   77 &   68.7&-&  225 &92.0 & - \\
CTransR \citep{AAAI159571}  &   75  & 70.2 &  -& 218 &  92.3& -  \\
KG2E \citep{He:2015}  &  59 & 74.0& -&  331 & 92.8 & - \\
TransD \citep{ji-EtAl:2015:ACL-IJCNLP}  &   91 & {77.3}&  -&  212 &  92.2 & - \\
lppTransD \citep{yoon-EtAl:2016:N16-1} & 78  & 78.7 & -& 270 & 94.3 & - \\
TransG \citep{xiao-huang-zhu:2016:P16-1} & 98 & 79.8& -& 470 & 93.3 & - \\
TranSparse \citep{JiLH016} & 82  & 79.5& -& {211}  & 93.2 & - \\
TranSparse-DT \citep{8057770} & 79 & 80.2 & -& 221 & 94.3 & - \\
ITransF \citep{xie-EtAl:2017:Long} & 65 & 81.0 & -& \underline{205} & 94.2 & - \\
NTN \citep{NIPS2013_5028} [$\blacklozenge$] & - & 41.4& 0.25& - & {66.1}  & 0.53 \\
TransE \citep{NIPS2013_5071} [$\blacksquare$] & - & 74.9 & 0.463& -& 94.3 & 0.495  \\
HolE \citep{Nickel:2016:HEK:3016100.3016172} & - & 73.9 & 0.524& -& {94.9} & {0.938}  \\
ComplEx \citep{TrouillonWRGB16} & - & {84.0} & {0.692}& - & 94.7 & {0.941}   \\
ANALOGY \citep{pmlr-v70-liu17d}& - & 85.4 & 0.725& - & 94.7 & {0.942}   \\
SimplE \citep{NIPS2018_7682} & - &  83.8 & 0.727 & - &  94.7 & 0.942 \\
TorusE \citep{Ebisu2018} & - & {83.2} & {0.733}& - & {95.4} & {0.947}  \\
STransE \citep{NguyenNAACL2016}  & 69  & {79.7}& {0.543}&  {206}  & {93.4}& 0.657  \\
ER-MLP \citep{Dong:2014} [$\spadesuit$] & 81 & 80.1 & 0.570& 299 & 94.2 & 0.895  \\
DISTMULT \citep{yang-etal-2015} [$\clubsuit$] & {42} & {89.3} & \underline{0.798} & 655  & 94.6 &  0.797  \\
ConvE \citep{DettmersMSR17} & 64&87.3 & {0.745}&504 & {95.5} & {0.942}   \\
 
  HypER \cite{balazevic2019hypernetwork} & 44 & 88.5 & 0.790 & 431 & 95.8 & \underline{0.951} \\
  RotatE \citep{sun2018rotate} & 40 & 88.4 & 0.797 & 309 & \underline{95.9} & {0.949} \\
 
  QuatE \citep{NIPS2019_8541} & \textbf{17} & 90.0 & 0.782 & \textbf{162} &\underline{95.9}& 0.950
\\
 
ComplEx-N3 \citep{LacroixUO18} & - & \underline{91} & \textbf{0.86} & -& \textbf{96} & {0.95}  \\

TuckER \citep{balazevic-etal-2019-tucker} & - & 89.2 & 0.795 & - & 95.8 & \textbf{0.953}\\

IRN \citep{shen-EtAl:2017:RepL4NLP1} & {38} & \textbf{92.7} & -& 249 & {95.3} & -   \\
ProjE \citep{ShiW16a} & \underline{34} & {88.4} & -& - & - & -  \\
\hline
\hline 
\textsc{r}TransE \citep{garciaduran-bordes-usunier:2015:EMNLP}  & {{50}} & 76.2  & -&  -& - & -  \\
PTransE-ADD \citep{lin-EtAl:2015:EMNLP1}  & {58} & {84.6}& -& - & -  & -  \\
PTransE-RNN \citep{lin-EtAl:2015:EMNLP1}  & {92} & {82.2}& -& - & -  & -  \\
GAKE \citep{feng-EtAl:2016:COLING1} & 119 & 64.8& -& - & - & -  \\
Gaifman \citep{NIPS2016_6098} & 75 & 84.2 & -& 352  & 93.9 & -  \\
Hiri \citep{Liu:2016:HRW:2911451.2911509} & -  & 70.3& 0.603& -  & 90.8 & 0.691 \\
Neural LP \citep{NIPS2017_6826}& - & 83.7 & {0.76}& - & 94.5 & \textbf{0.94}   \\
R-GCN+ \citep{SchlichtkrullKB17} & - & 84.2 & 0.696& - & \textbf{96.4} & {0.819}  \\
KB$_{LRN}$ \citep{Alberto17} &  \textbf{44} &  \textbf{87.5} & \textbf{0.794}& - & - & -  \\
\hline
TEKE\_H  \citep{DBLP:conf/ijcai/WangL16} & 108 & 73.0& -&  \textbf{114}  & 92.9 & -  \\
SSP \citep{0005HZ16} & 82 & 79.0 & -& 156 & 93.2 & -  \\

\hline
\end{tabular}
}
\caption{Entity prediction results  on  WN18 and FB15k, which are taken from the corresponding papers. \textbf{MR} and \textbf{@10} denote metrics mean rank and Hits@10 (in \%), respectively. 
[$\blacklozenge$], \textbf{[$\blacksquare$]},  \textbf{[$\spadesuit$]} and \textbf{[$\clubsuit$]} denote results taking from Yang {et al.} (2015), Nickel {et al.} (2016b), Ravishankar {et al.} (2017) and Kadlec {et al.} (2017), 
respectively.}
\label{02tab:linkprediction}
\end{table}

\begin{table}[!t]
\centering
\resizebox{10cm}{!}{
\def\arraystretch{0.95}
\begin{tabular}{l|lll|lll}
\hline
\multirow{3}{*}{\bf Method}& \multicolumn{6}{|c}{\textbf{Filtered}}\\
\cline{2-7}
& \multicolumn{3}{|c}{\bf FB15k-237} & \multicolumn{3}{|c}{\bf WN18RR} \\
\cline{2-7}
 &   MR & @10 & MRR&   MR & @10 & MRR \\
 \hline
IRN \citep{shen-EtAl:2017:RepL4NLP1} & {211} & 46.4 & - & - & - & -\\
KBGAN \citep{Cai2017} & - & 45.8 & 0.278 & - & 48.1 & 0.213 \\
DISTMULT \citep{yang-etal-2015} [$\blacklozenge$] & 254 & 41.9  & 0.241  & 5110   & 49 & {0.43} \\
ComplEx \citep{TrouillonWRGB16} [$\blacklozenge$] & 339 & 42.8 & 0.247 & 5261 & {51} & {0.44}\\
ConvE \citep{DettmersMSR17} & 246 & 49.1 &  {0.316} & 5277 & 48 & {0.46} \\
ER-MLP \citep{Dong:2014} [$\spadesuit$]  & {219} &  {54.0} & {0.342} & {4798} &  41.9 & {0.366}\\

 HypER \cite{balazevic2019hypernetwork} & 250 & 52.0 & 0.341 & 5798 & 52.2 & 0.465 \\ 

TransE \citep{NIPS2013_5071} [$\blacksquare$] & 347 & 46.5 & 0.294 & \underline{743} &  {56.0} & 0.245\\

ConvKB \citep{NguyenNNP2017} [$\blacksquare$] & 254 &  53.2 & \underline{0.418} & 763 & 56.7 & 0.253 \\
 CapsE \citep{NguyenVNNP_NAACL2019}  & 303 & \textbf{59.3} & \textbf{0.523} & \textbf{719} & 56.0 & 0.415\\

  InteractE  \citep{interacte2020} &  \underline{172} & 53.5 & 0.354 &  5202 & 52.8 & 0.463\\
  
  RotatE \citep{sun2018rotate} & {177} & 53.3  & 0.338 & {3340} & \underline{57.1} & {0.476}\\

  QuatE \citep{NIPS2019_8541} & \textbf{87} &  55.0 & 0.348 & 2314 & \textbf{58.2} & \textbf{0.488}\\

 ComplEx-N3 \citep{LacroixUO18} & - & \underline{56} & {0.37} & - & {57} & \underline{0.48}\\
Conv-TransE \citep{Shang19} & - & 51 & 0.33 & - & 52 & 0.46 \\

TuckER \citep{balazevic-etal-2019-tucker} & - & 54.4 & 0.358 & - & 52.6 & 0.470\\
\hline 
\hline
Neural LP \citep{NIPS2017_6826}&  - & 36.2 & 0.24 & - & - & -\\
R-GCN+ \citep{SchlichtkrullKB17} & - & 41.7 & 0.249 & - & - & -  \\
 KB$_{LRN}$ \citep{Alberto17} & {209} & {49.3} & {0.309}& - & - & -\\
KBGAT \citep{nathani-etal-2019-learning} & 210 & \textbf{62.6} & \textbf{0.518} & {1940} & {58.1} & 0.440 \\
ReInceptionE \cite{xie-etal-2020-reinceptione} & \textbf{173} & 52.8 & 0.349 & \textbf{1894} &  \textbf{58.2} & \textbf{0.483} \\ 
SACN \citep{Shang19} & - & {54} & {0.35} & - & {54} & {0.47} \\
\hline
\end{tabular}
}
\caption{Entity prediction results  on  WN18RR and FB15k-237, which are taken from the corresponding papers.\ 
\textbf{[$\blacklozenge$]}, \textbf{[$\spadesuit$]} and \textbf{[$\blacksquare$]} denote results taking from Dettmers {et al.} (2018), Ravishankar {et al.} (2017) and Nguyen {et al.} (2019), 
 respectively.
}
\label{tab:linkprediction2}
\end{table}

\subsection{Main Results}

Tables \ref{02tab:linkprediction} and \ref{tab:linkprediction2}  list  recent  entity prediction results of KG completion models on FB15k and WN18 and on FB15k-237 and WN18RR, respectively. In Table \ref{02tab:linkprediction},  the first 28 rows report the performance of  triple-based models that directly optimize a score function for the triples in a KG, i.e. they 
do not exploit information about alternative paths between head and tail entities. The next 9 rows report results of   models that exploit information about relation paths or neighborhood information. The last 2 rows present results for  models  which make  use of textual mentions derived from a large external corpus.  In Table \ref{tab:linkprediction2}, the last 5 rows report results of  models that exploit the path or neighborhood information.

In general, Tables \ref{02tab:linkprediction}  and  \ref{tab:linkprediction2} show that the models using  external corpus
information or  employing path information achieve better scores
than the triple-based models that do not use such information. 
In terms of models not exploiting path 
or external information, the  complex vector-based models (e.g. QuatE, CompleEx-N3 and RotatE) produce the strongest evaluation scores, followed by the neural network-based models (e.g. CapsE, InteractE and HypER).\footnote{CapsE  uses the pre-trained  word embeddings for entity vector initialization on WN18RR.  It is not surprising that CapsE produces the best MR on WN18RR as many entity names in WordNet are lexically meaningful. It is possible for all other embedding models to utilize the pre-trained word vectors as well. However, averaging the pre-trained word embeddings for initializing entity vectors  is an open problem, and it is not always useful since entity names in many domain-specific KGs are not lexically meaningful \citep{AAAI148531,guu-miller-liang:2015:EMNLP}.}  
Tables \ref{02tab:linkprediction}  and  \ref{tab:linkprediction2}  also show that TransE and DISTMULT, despite of theirs  simplicity, can produce very competitive results  (i.e. by performing a careful grid search of hyper-parameters). 

\section{Discussion and Conclusion}

The reasons why much work has been devoted towards developing triple-based models are: (1) additional information sources might not be available, e.g., for KGs for specialized
domains, (2) models that do not exploit path information or external
resources are simpler and thus typically much faster
to train than the more complex models using path or external 
information, and (3) the more complex models
that exploit path or external information are typically
extensions of these simpler models, and are often
initialized with parameters estimated by such simpler
models, so improvements to the simpler models
should yield corresponding improvements to the
more complex models as well \cite{NguyenNAACL2016}.

It is worth to further explore those KG completion embedding models for a new application  where we could formulate its corresponding data into triples. For example, in   Web search  engines, we observe {user}-oriented relationships between submitted {queries} and  {documents} returned by the search  engines.  That is, we have triple representations (query, user, document) in which  for each user-oriented relationship, we would have many queries and documents, resulting in a lot of Many-to-Many relationships. Inspired by this observation, 
\newcite{NguyenECIR2017} applied  STransE  \citep{NguyenNAACL2016} for  {search personalization} to re-rank the search documents returned by a search engine for users' submitted  queries. Other application examples can be also found  for recommender  systems  \citep{10.1145/2939672.2939673,10.1145/3109859.3109882,10.1145/3308558.3313705}, social relation extraction \citep{ijcai2017-399} and visual relation detection \citep{Zhang_2017_CVPR}.  

Future research  directions might also include: (i) Combining logical rules which contain  rich background information and KG triples in a unified KG completion framework, e.g. jointly embedding KGs and logical rules  \citep{guo-etal-2016-jointly,NIPS2017_6826}. (ii) Recent embedding models for KG completion hold a closed-world assumption where the KGs are fixed (i.e. new entities might not be  added easily), therefore it would be worth exploring open-world KG completion models to connect unseen entities to the existing KGs \citep{Shi2018}. (iii) Investigating efficient approaches which can be applied to large-scale KGs of millions of entities and relations \cite{Zhang2020Efficient}.

In this paper, we have presented a comprehensive survey of embedding models of entity and relationships for knowledge graph completion. This paper also provides update-to-date experimental results of the embedding models for the entity prediction (i.e. link prediction) task  on  benchmark datasets FB15k, WN18, FB15k-237 and WN18RR. 
We hope that this paper serves its purpose by providing a concrete foundation for future research and applications on the topic.


\bibliographystyle{coling}
\bibliography{REFs}

\begin{thebibliography}{}

\bibitem[\protect\citename{Agirre \bgroup et al.\egroup }2013]{Agirre2013}
Eneko Agirre, Oier L{\'o}pez~de Lacalle, and Aitor Soroa.
\newblock 2013.
\newblock {Random Walks for Knowledge-Based Word Sense Disambiguation}.
\newblock {\em Computational Linguistics}, 40(1):57--84.

\bibitem[\protect\citename{Baeza{-}Yates and Ribeiro{-}Neto}2011]{Ricardo2011}
Ricardo~A. Baeza{-}Yates and Berthier~A. Ribeiro{-}Neto.
\newblock 2011.
\newblock {\em Modern Information Retrieval - the concepts and technology
  behind search, Second edition}.
\newblock Pearson Education Ltd., Harlow, England.

\bibitem[\protect\citename{Bala\v{z}evi\'c \bgroup et al.\egroup
  }2019]{balazevic2019hypernetwork}
Ivana Bala\v{z}evi\'c, Carl Allen, and Timothy~M Hospedales.
\newblock 2019.
\newblock Hypernetwork knowledge graph embeddings.
\newblock In {\em ICANN}, pages 553--565.

\bibitem[\protect\citename{Balazevic \bgroup et al.\egroup
  }2019]{balazevic-etal-2019-tucker}
Ivana Balazevic, Carl Allen, and Timothy Hospedales.
\newblock 2019.
\newblock {T}uck{ER}: Tensor factorization for knowledge graph completion.
\newblock In {\em EMNLP-IJCNLP}, pages 5185--5194.

\bibitem[\protect\citename{Berant \bgroup et al.\egroup
  }2013]{berant-EtAl:2013:EMNLP}
Jonathan Berant, Andrew Chou, Roy Frostig, and Percy Liang.
\newblock 2013.
\newblock {Semantic Parsing on {Freebase} from Question-Answer Pairs}.
\newblock In {\em EMNLP}, pages 1533--1544.

\bibitem[\protect\citename{Bollacker \bgroup et al.\egroup
  }2008]{Bollacker:2008}
Kurt Bollacker, Colin Evans, Praveen Paritosh, Tim Sturge, and Jamie Taylor.
\newblock 2008.
\newblock {Freebase: A Collaboratively Created Graph Database for Structuring
  Human Knowledge}.
\newblock In {\em SIGMOD}, pages 1247--1250.

\bibitem[\protect\citename{Bordes \bgroup et al.\egroup }2011]{bordes-2011}
Antoine Bordes, Jason Weston, Ronan Collobert, and Yoshua Bengio.
\newblock 2011.
\newblock {Learning Structured Embeddings of Knowledge Bases}.
\newblock In {\em AAAI}, pages 301--306.

\bibitem[\protect\citename{Bordes \bgroup et al.\egroup }2012]{Bordes2014SME}
Antoine Bordes, Xavier Glorot, Jason Weston, and Yoshua Bengio.
\newblock 2012.
\newblock {A Semantic Matching Energy Function for Learning with
  Multi-relational Data}.
\newblock {\em Machine Learning}, 94(2):233--259.

\bibitem[\protect\citename{Bordes \bgroup et al.\egroup }2013]{NIPS2013_5071}
Antoine Bordes, Nicolas Usunier, Alberto Garcia-Duran, Jason Weston, and Oksana
  Yakhnenko.
\newblock 2013.
\newblock {Translating Embeddings for Modeling Multi-relational Data}.
\newblock In {\em NIPS}, pages 2787--2795.

\bibitem[\protect\citename{Cai and Wang}2018]{Cai2017}
Liwei Cai and William~Yang Wang.
\newblock 2018.
\newblock {KBGAN: Adversarial Learning for Knowledge Graph Embeddings}.
\newblock In {\em NAACL-HLT}, pages 1470--1480.

\bibitem[\protect\citename{Cao \bgroup et al.\egroup
  }2019]{10.1145/3308558.3313705}
Yixin Cao, Xiang Wang, Xiangnan He, Zikun Hu, and Tat-Seng Chua.
\newblock 2019.
\newblock Unifying knowledge graph learning and recommendation: Towards a
  better understanding of user preferences.
\newblock In {\em WWW}, page 151–161.

\bibitem[\protect\citename{Carlson \bgroup et al.\egroup
  }2010]{Carlson:2010:TAN:2898607.2898816}
Andrew Carlson, Justin Betteridge, Bryan Kisiel, Burr Settles, Estevam~R.
  Hruschka, Jr., and Tom~M. Mitchell.
\newblock 2010.
\newblock {Toward an Architecture for Never-ending Language Learning}.
\newblock In {\em AAAI}, pages 1306--1313.

\bibitem[\protect\citename{Chang \bgroup et al.\egroup }2017]{8057770}
L.~Chang, M.~Zhu, T.~Gu, C.~Bin, J.~Qian, and J.~Zhang.
\newblock 2017.
\newblock {Knowledge Graph Embedding by Dynamic Translation}.
\newblock {\em IEEE Access}, 5:20898--20907.

\bibitem[\protect\citename{Chen \bgroup et al.\egroup }2018]{N18-1165}
Wenhu Chen, Wenhan Xiong, Xifeng Yan, and William~Yang Wang.
\newblock 2018.
\newblock Variational knowledge graph reasoning.
\newblock In {\em NAACL-HLT}, pages 1823--1832.

\bibitem[\protect\citename{Das \bgroup et al.\egroup
  }2017]{das-EtAl:2017:EACLlong1}
Rajarshi Das, Arvind Neelakantan, David Belanger, and Andrew McCallum.
\newblock 2017.
\newblock {Chains of Reasoning over Entities, Relations, and Text using
  Recurrent Neural Networks}.
\newblock In {\em EACL}, pages 132--141.

\bibitem[\protect\citename{Das \bgroup et al.\egroup }2018]{das2018go}
Rajarshi Das, Shehzaad Dhuliawala, Manzil Zaheer, Luke Vilnis, Ishan Durugkar,
  Akshay Krishnamurthy, Alex Smola, and Andrew McCallum.
\newblock 2018.
\newblock {Go for a Walk and Arrive at the Answer: Reasoning Over Paths in
  Knowledge Bases using Reinforcement Learning}.
\newblock In {\em ICLR}.

\bibitem[\protect\citename{Dettmers \bgroup et al.\egroup }2018]{DettmersMSR17}
Tim Dettmers, Pasquale Minervini, Pontus Stenetorp, and Sebastian Riedel.
\newblock 2018.
\newblock {Convolutional 2D Knowledge Graph Embeddings}.
\newblock In {\em AAAI}, pages 1811--1818.

\bibitem[\protect\citename{Devlin \bgroup et al.\egroup
  }2019]{devlin-etal-2019-bert}
Jacob Devlin, Ming-Wei Chang, Kenton Lee, and Kristina Toutanova.
\newblock 2019.
\newblock {BERT}: Pre-training of deep bidirectional transformers for language
  understanding.
\newblock In {\em NAACL-HLT}, pages 4171--4186.

\bibitem[\protect\citename{Dong \bgroup et al.\egroup }2014]{Dong:2014}
Xin Dong, Evgeniy Gabrilovich, Geremy Heitz, Wilko Horn, Ni~Lao, Kevin Murphy,
  Thomas Strohmann, Shaohua Sun, and Wei Zhang.
\newblock 2014.
\newblock Knowledge vault: A web-scale approach to probabilistic knowledge
  fusion.
\newblock In {\em KDD}, pages 601--610.

\bibitem[\protect\citename{Dur{\'{a}}n and Niepert}2018]{Alberto17}
Alberto~Garc{\'{\i}}a Dur{\'{a}}n and Mathias Niepert.
\newblock 2018.
\newblock {KBLRN}: End-to-end learning of knowledge base representations with
  latent, relational, and numerical features.
\newblock In {\em UAI}.

\bibitem[\protect\citename{Dutta and Weikum}2015]{TACL522}
Sourav Dutta and Gerhard Weikum.
\newblock 2015.
\newblock {Cross-Document Co-Reference Resolution using Sample-Based Clustering
  with Knowledge Enrichment}.
\newblock {\em Transactions of ACL}, 3:15--28.

\bibitem[\protect\citename{Ebisu and Ichise}2018]{Ebisu2018}
Takuma Ebisu and Ryutaro Ichise.
\newblock 2018.
\newblock {TorusE: Knowledge Graph Embedding on a Lie Group}.
\newblock In {\em AAAI}, pages 1819--1826.

\bibitem[\protect\citename{Fader \bgroup et al.\egroup
  }2014]{Fader:2014:OQA:2623330.2623677}
Anthony Fader, Luke Zettlemoyer, and Oren Etzioni.
\newblock 2014.
\newblock {Open Question Answering over Curated and Extracted Knowledge Bases}.
\newblock In {\em KDD}, pages 1156--1165.

\bibitem[\protect\citename{Feldman \bgroup et al.\egroup
  }2019]{commonsense2019}
Joshua Feldman, Joe Davison, and Alexander~M. Rush.
\newblock 2019.
\newblock {Commonsense Knowledge Mining from Pretrained Models}.
\newblock In {\em EMNLP-IJCNLP}, pages 1173--1178.

\bibitem[\protect\citename{Fellbaum}1998]{FellbaumC98}
Christiane~D. Fellbaum.
\newblock 1998.
\newblock {\em {WordNet}: An Electronic Lexical Database}.
\newblock MIT Press.

\bibitem[\protect\citename{Feng \bgroup et al.\egroup }2016a]{FengHWZHZ16}
Jun Feng, Minlie Huang, Mingdong Wang, Mantong Zhou, Yu~Hao, and Xiaoyan Zhu.
\newblock 2016a.
\newblock Knowledge graph embedding by flexible translation.
\newblock In {\em KR}, pages 557--560.

\bibitem[\protect\citename{Feng \bgroup et al.\egroup
  }2016b]{feng-EtAl:2016:COLING1}
Jun Feng, Minlie Huang, Yang Yang, and xiaoyan zhu.
\newblock 2016b.
\newblock {GAKE: Graph Aware Knowledge Embedding}.
\newblock In {\em COLING}, pages 641--651.

\bibitem[\protect\citename{Ferrucci}2012]{Ferrucci:2012:ITW:2481742.2481743}
David~Angelo Ferrucci.
\newblock 2012.
\newblock {Introduction to "This is Watson"}.
\newblock {\em IBM Journal of Research and Development}, 56(3):235--249.

\bibitem[\protect\citename{Garc{\'{\i}}a{-}Dur{\'{a}}n \bgroup et al.\egroup
  }2015]{garciaduran-bordes-usunier:2015:EMNLP}
Alberto Garc{\'{\i}}a{-}Dur{\'{a}}n, Antoine Bordes, and Nicolas Usunier.
\newblock 2015.
\newblock {Composing Relationships with Translations}.
\newblock In {\em EMNLP}, pages 286--290.

\bibitem[\protect\citename{Garc{\'{\i}}a{-}Dur{\'{a}}n \bgroup et al.\egroup
  }2016]{Garcia-DuranBUG15}
Alberto Garc{\'{\i}}a{-}Dur{\'{a}}n, Antoine Bordes, Nicolas Usunier, and Yves
  Grandvalet.
\newblock 2016.
\newblock {Combining Two and Three-Way Embedding Models for Link Prediction in
  Knowledge Bases}.
\newblock {\em Journal of Artificial Intelligence Research}, 55:715--742.

\bibitem[\protect\citename{Gardner and
  Mitchell}2015]{gardner-mitchell:2015:EMNLP}
Matt Gardner and Tom Mitchell.
\newblock 2015.
\newblock {Efficient and Expressive Knowledge Base Completion Using Subgraph
  Feature Extraction}.
\newblock In {\em EMNLP}, pages 1488--1498.

\bibitem[\protect\citename{Gardner \bgroup et al.\egroup }2014]{Gardner2014}
Matt Gardner, Partha~P. Talukdar, Jayant Krishnamurthy, and Tom~M. Mitchell.
\newblock 2014.
\newblock {Incorporating Vector Space Similarity in Random Walk Inference over
  Knowledge Bases}.
\newblock In {\em EMNLP}, pages 397--406.

\bibitem[\protect\citename{Guo \bgroup et al.\egroup
  }2016]{guo-etal-2016-jointly}
Shu Guo, Quan Wang, Lihong Wang, Bin Wang, and Li~Guo.
\newblock 2016.
\newblock Jointly embedding knowledge graphs and logical rules.
\newblock In {\em EMNLP}, pages 192--202.

\bibitem[\protect\citename{Guu \bgroup et al.\egroup
  }2015]{guu-miller-liang:2015:EMNLP}
Kelvin Guu, John Miller, and Percy Liang.
\newblock 2015.
\newblock {Traversing Knowledge Graphs in Vector Space}.
\newblock In {\em EMNLP}, pages 318--327.

\bibitem[\protect\citename{He \bgroup et al.\egroup }2015]{He:2015}
Shizhu He, Kang Liu, Guoliang Ji, and Jun Zhao.
\newblock 2015.
\newblock {Learning to Represent Knowledge Graphs with Gaussian Embedding}.
\newblock In {\em CIKM}, pages 623--632.

\bibitem[\protect\citename{He \bgroup et al.\egroup
  }2017]{10.1145/3109859.3109882}
Ruining He, Wang-Cheng Kang, and Julian McAuley.
\newblock 2017.
\newblock Translation-based recommendation.
\newblock In {\em RecSys}, page 161–169.

\bibitem[\protect\citename{Jenatton \bgroup et al.\egroup }2012]{NIPS2012_4744}
Rodolphe Jenatton, Nicolas~L. Roux, Antoine Bordes, and Guillaume~R Obozinski.
\newblock 2012.
\newblock {A latent factor model for highly multi-relational data}.
\newblock In {\em NIPS}, pages 3167--3175.

\bibitem[\protect\citename{Ji \bgroup et al.\egroup
  }2015]{ji-EtAl:2015:ACL-IJCNLP}
Guoliang Ji, Shizhu He, Liheng Xu, Kang Liu, and Jun Zhao.
\newblock 2015.
\newblock {Knowledge Graph Embedding via Dynamic Mapping Matrix}.
\newblock In {\em ACL-IJCNLP}, pages 687--696.

\bibitem[\protect\citename{Ji \bgroup et al.\egroup }2016]{JiLH016}
Guoliang Ji, Kang Liu, Shizhu He, and Jun Zhao.
\newblock 2016.
\newblock {Knowledge Graph Completion with Adaptive Sparse Transfer Matrix}.
\newblock In {\em AAAI}, pages 985--991.

\bibitem[\protect\citename{Kazemi and Poole}2018]{NIPS2018_7682}
Seyed~Mehran Kazemi and David Poole.
\newblock 2018.
\newblock Simple embedding for link prediction in knowledge graphs.
\newblock In {\em NIPS}, pages 4284--4295.

\bibitem[\protect\citename{Kipf and Welling}2017]{KipfW17}
Thomas~N. Kipf and Max Welling.
\newblock 2017.
\newblock {Semi-Supervised Classification with Graph Convolutional Networks}.
\newblock In {\em ICLR}.

\bibitem[\protect\citename{Krishnamurthy and
  Mitchell}2012]{krishnamurthy-mitchell:2012:EMNLP-CoNLL}
Jayant Krishnamurthy and Tom Mitchell.
\newblock 2012.
\newblock {Weakly Supervised Training of Semantic Parsers}.
\newblock In {\em EMNLP-CoNLL}, pages 754--765.

\bibitem[\protect\citename{Krompa{\ss} \bgroup et al.\egroup }2015]{Denis2015}
Denis Krompa{\ss}, Stephan Baier, and Volker Tresp.
\newblock 2015.
\newblock {Type-Constrained Representation Learning in Knowledge Graphs}.
\newblock In {\em ISWC}, pages 640--655.

\bibitem[\protect\citename{Lacroix \bgroup et al.\egroup }2018]{LacroixUO18}
Timoth{\'{e}}e Lacroix, Nicolas Usunier, and Guillaume Obozinski.
\newblock 2018.
\newblock {Canonical Tensor Decomposition for Knowledge Base Completion}.
\newblock In {\em ICML}, pages 2869--2878.

\bibitem[\protect\citename{Lao and Cohen}2010]{Lao2010}
Ni~Lao and William~W. Cohen.
\newblock 2010.
\newblock {Relational retrieval using a combination of path-constrained random
  walks}.
\newblock {\em Machine Learning}, 81(1):53--67.

\bibitem[\protect\citename{Lao \bgroup et al.\egroup
  }2011]{Lao:2011:RWI:2145432.2145494}
Ni~Lao, Tom Mitchell, and William~W. Cohen.
\newblock 2011.
\newblock {Random Walk Inference and Learning in a Large Scale Knowledge Base}.
\newblock In {\em EMNLP}, pages 529--539.

\bibitem[\protect\citename{Lehmann \bgroup et al.\egroup
  }2015]{LehmannIJJKMHMK15}
Jens Lehmann, Robert Isele, Max Jakob, Anja Jentzsch, Dimitris Kontokostas,
  Pablo~N. Mendes, Sebastian Hellmann, Mohamed Morsey, Patrick van Kleef,
  S{\"{o}}ren Auer, and Christian Bizer.
\newblock 2015.
\newblock {DBpedia - A Large-scale, Multilingual Knowledge Base Extracted from
  Wikipedia}.
\newblock {\em Semantic Web}, 6(2):167--195.

\bibitem[\protect\citename{Liang and Forbus}2015]{LiangF15}
Chen Liang and Kenneth~D. Forbus.
\newblock 2015.
\newblock {Learning Plausible Inferences from Semantic Web Knowledge by
  Combining Analogical Generalization with Structured Logistic Regression}.
\newblock In {\em AAAI}, pages 551--557.

\bibitem[\protect\citename{Lin \bgroup et al.\egroup
  }2015a]{lin-EtAl:2015:EMNLP1}
Yankai Lin, Zhiyuan Liu, Huanbo Luan, Maosong Sun, Siwei Rao, and Song Liu.
\newblock 2015a.
\newblock {Modeling Relation Paths for Representation Learning of Knowledge
  Bases}.
\newblock In {\em EMNLP}, pages 705--714.

\bibitem[\protect\citename{Lin \bgroup et al.\egroup }2015b]{AAAI159571}
Yankai Lin, Zhiyuan Liu, Maosong Sun, Yang Liu, and Xuan Zhu.
\newblock 2015b.
\newblock {Learning Entity and Relation Embeddings for Knowledge Graph
  Completion}.
\newblock In {\em AAAI}, pages 2181--2187.

\bibitem[\protect\citename{Liu \bgroup et al.\egroup
  }2016]{Liu:2016:HRW:2911451.2911509}
Qiao Liu, Liuyi Jiang, Minghao Han, Yao Liu, and Zhiguang Qin.
\newblock 2016.
\newblock {Hierarchical Random Walk Inference in Knowledge Graphs}.
\newblock In {\em SIGIR}, pages 445--454.

\bibitem[\protect\citename{Liu \bgroup et al.\egroup }2017]{pmlr-v70-liu17d}
Hanxiao Liu, Yuexin Wu, and Yiming Yang.
\newblock 2017.
\newblock {Analogical Inference for Multi-relational Embeddings}.
\newblock In {\em ICML}, pages 2168--2178.

\bibitem[\protect\citename{Luo \bgroup et al.\egroup
  }2015]{luo-EtAl:2015:EMNLP3}
Yuanfei Luo, Quan Wang, Bin Wang, and Li~Guo.
\newblock 2015.
\newblock {Context-Dependent Knowledge Graph Embedding}.
\newblock In {\em EMNLP}, pages 1656--1661.

\bibitem[\protect\citename{Mazumder and Liu}2017]{ijcai2017-166}
Sahisnu Mazumder and Bing Liu.
\newblock 2017.
\newblock {Context-aware Path Ranking for Knowledge Base Completion}.
\newblock In {\em IJCAI}, pages 1195--1201.

\bibitem[\protect\citename{Mikolov \bgroup et al.\egroup
  }2013]{MikolovNIPS2013}
Tomas Mikolov, Ilya Sutskever, Kai Chen, Greg~S Corrado, and Jeff Dean.
\newblock 2013.
\newblock {Distributed Representations of Words and Phrases and their
  Compositionality}.
\newblock In {\em NIPS}, pages 3111--3119.

\bibitem[\protect\citename{Miller}1995]{Miller:1995:WLD:219717.219748}
George~A. Miller.
\newblock 1995.
\newblock {WordNet: A Lexical Database for English}.
\newblock {\em Communications of the ACM}, 38(11):39--41.

\bibitem[\protect\citename{Nathani \bgroup et al.\egroup
  }2019]{nathani-etal-2019-learning}
Deepak Nathani, Jatin Chauhan, Charu Sharma, and Manohar Kaul.
\newblock 2019.
\newblock Learning attention-based embeddings for relation prediction in
  knowledge graphs.
\newblock In {\em ACL}, pages 4710--4723.

\bibitem[\protect\citename{Navigli and
  Velardi}2005]{Navigli:2005:SSI:1070615.1070793}
Roberto Navigli and Paola Velardi.
\newblock 2005.
\newblock {Structural Semantic Interconnections: A Knowledge-Based Approach to
  Word Sense Disambiguation}.
\newblock {\em IEEE Transactions on Pattern Analysis and Machine Intelligence},
  27(7):1075--1086.

\bibitem[\protect\citename{Neelakantan \bgroup et al.\egroup
  }2015]{neelakantan-roth-mccallum:2015:ACL-IJCNLP}
Arvind Neelakantan, Benjamin Roth, and Andrew McCallum.
\newblock 2015.
\newblock {Compositional Vector Space Models for Knowledge Base Completion}.
\newblock In {\em ACL-IJCNLP}, pages 156--166.

\bibitem[\protect\citename{Nguyen \bgroup et al.\egroup
  }2016a]{NguyenNAACL2016}
Dat~Quoc Nguyen, Kairit Sirts, Lizhen Qu, and Mark Johnson.
\newblock 2016a.
\newblock {STransE: a novel embedding model of entities and relationships in
  knowledge bases}.
\newblock In {\em NAACL-HLT}, pages 460--466.

\bibitem[\protect\citename{Nguyen \bgroup et al.\egroup
  }2016b]{NguyenCoNLL2016}
Dat~Quoc Nguyen, Kairit Sirts, Lizhen Qu, and Mark Johnson.
\newblock 2016b.
\newblock {Neighborhood Mixture Model for Knowledge Base Completion}.
\newblock In {\em CoNLL}, pages 40--50.

\bibitem[\protect\citename{Nguyen \bgroup et al.\egroup }2018]{NguyenNNP2017}
Dai~Quoc Nguyen, Tu~Dinh Nguyen, Dat~Quoc Nguyen, and Dinh Phung.
\newblock 2018.
\newblock {A Novel Embedding Model for Knowledge Base Completion Based on
  Convolutional Neural Network}.
\newblock In {\em NAACL-HLT}, pages 327--333.

\bibitem[\protect\citename{Nguyen \bgroup et al.\egroup }2019a]{SWJ318}
Dai~Quoc Nguyen, Dat~Quoc Nguyen, Tu~Dinh Nguyen, and Dinh Phung.
\newblock 2019a.
\newblock {A convolutional neural network-based model for knowledge base
  completion and its application to search personalization}.
\newblock {\em Semantic Web}, 10(5).

\bibitem[\protect\citename{Nguyen \bgroup et al.\egroup
  }2019b]{NguyenVNNP_NAACL2019}
Dai~Quoc Nguyen, Thanh Vu, Tu~Dinh Nguyen, Dat~Quoc Nguyen, and Dinh Phung.
\newblock 2019b.
\newblock {A Capsule Network-based Embedding Model for Knowledge Graph
  Completion and Search Personalization}.
\newblock In {\em NAACL-HLT}, pages pages 2180--2189.

\bibitem[\protect\citename{Nickel \bgroup et al.\egroup
  }2011]{ICML2011Nickel_438}
Maximilian Nickel, Volker Tresp, and Hans-Peter Kriegel.
\newblock 2011.
\newblock {A Three-Way Model for Collective Learning on Multi-Relational Data}.
\newblock In {\em ICML}, pages 809--816.

\bibitem[\protect\citename{Nickel \bgroup et al.\egroup }2016a]{NickelMTG15}
Maximilian Nickel, Kevin Murphy, Volker Tresp, and Evgeniy Gabrilovich.
\newblock 2016a.
\newblock {A Review of Relational Machine Learning for Knowledge Graphs}.
\newblock {\em the IEEE}, 104(1):11--33.

\bibitem[\protect\citename{Nickel \bgroup et al.\egroup
  }2016b]{Nickel:2016:HEK:3016100.3016172}
Maximilian Nickel, Lorenzo Rosasco, and Tomaso Poggio.
\newblock 2016b.
\newblock Holographic embeddings of knowledge graphs.
\newblock In {\em AAAI}, pages 1955--1961.

\bibitem[\protect\citename{Niepert}2016]{NIPS2016_6098}
Mathias Niepert.
\newblock 2016.
\newblock {Discriminative Gaifman Models}.
\newblock In {\em NIPS}, pages 3405--3413.

\bibitem[\protect\citename{Pennington \bgroup et al.\egroup
  }2014]{Pennington14}
Jeffrey Pennington, Richard Socher, and Christopher Manning.
\newblock 2014.
\newblock {Glove: Global Vectors for Word Representation}.
\newblock In {\em EMNLP}, pages 1532--1543.

\bibitem[\protect\citename{Ponzetto and
  Strube}2006]{ponzetto-strube:2006:HLT-NAACL06-Main}
Simone~Paolo Ponzetto and Michael Strube.
\newblock 2006.
\newblock {Exploiting Semantic Role Labeling, WordNet and Wikipedia for
  Coreference Resolution}.
\newblock In {\em NAACL}, pages 192--199.

\bibitem[\protect\citename{Sabour \bgroup et al.\egroup }2017]{NIPS2017_6975}
Sara Sabour, Nicholas Frosst, and Geoffrey~E Hinton.
\newblock 2017.
\newblock {Dynamic Routing Between Capsules}.
\newblock In {\em NIPS}, pages 3856--3866.

\bibitem[\protect\citename{Schlichtkrull \bgroup et al.\egroup
  }2018]{SchlichtkrullKB17}
Michael~Sejr Schlichtkrull, Thomas~N. Kipf, Peter Bloem, Rianne van~den Berg,
  Ivan Titov, and Max Welling.
\newblock 2018.
\newblock {Modeling Relational Data with Graph Convolutional Networks}.
\newblock In {\em ESWC}, pages 593--607.

\bibitem[\protect\citename{Shang \bgroup et al.\egroup }2019]{Shang19}
Chao Shang, Yun Tang, Jing Huang, Jinbo Bi, Xiaodong He, and Bowen Zhou.
\newblock 2019.
\newblock {End-to-end Structure-Aware Convolutional Networks for Knowledge Base
  Completion}.
\newblock In {\em AAAI}, pages 3060--3067.

\bibitem[\protect\citename{Shen \bgroup et al.\egroup
  }2017]{shen-EtAl:2017:RepL4NLP1}
Yelong Shen, Po-Sen Huang, Ming-Wei Chang, and Jianfeng Gao.
\newblock 2017.
\newblock {Modeling Large-Scale Structured Relationships with Shared Memory for
  Knowledge Base Completion}.
\newblock In {\em Rep4NLP}, pages 57--68.

\bibitem[\protect\citename{Shi and Weninger}2017]{ShiW16a}
Baoxu Shi and Tim Weninger.
\newblock 2017.
\newblock {ProjE: Embedding Projection for Knowledge Graph Completion}.
\newblock In {\em AAAI}, pages 1236--1242.

\bibitem[\protect\citename{Shi and Weninger}2018]{Shi2018}
Baoxu Shi and Tim Weninger.
\newblock 2018.
\newblock {Open-world knowledge graph completion}.
\newblock In {\em AAAI}, pages 1957--1964.

\bibitem[\protect\citename{Socher \bgroup et al.\egroup }2013]{NIPS2013_5028}
Richard Socher, Danqi Chen, Christopher~D Manning, and Andrew Ng.
\newblock 2013.
\newblock {Reasoning With Neural Tensor Networks for Knowledge Base
  Completion}.
\newblock In {\em NIPS}, pages 926--934.

\bibitem[\protect\citename{Suchanek \bgroup et al.\egroup }2007]{Suchanek:2007}
Fabian~M. Suchanek, Gjergji Kasneci, and Gerhard Weikum.
\newblock 2007.
\newblock {YAGO: A Core of Semantic Knowledge}.
\newblock In {\em WWW}, pages 697--706.

\bibitem[\protect\citename{Sun \bgroup et al.\egroup }2019]{sun2018rotate}
Zhiqing Sun, Zhi-Hong Deng, Jian-Yun Nie, and Jian Tang.
\newblock 2019.
\newblock {RotatE: Knowledge Graph Embedding by Relational Rotation in Complex
  Space}.
\newblock In {\em ICLR}.

\bibitem[\protect\citename{{Szegedy} \bgroup et al.\egroup }2016]{7780677}
C.~{Szegedy}, V.~{Vanhoucke}, S.~{Ioffe}, J.~{Shlens}, and Z.~{Wojna}.
\newblock 2016.
\newblock {Rethinking the Inception Architecture for Computer Vision}.
\newblock In {\em CVPR}, pages 2818--2826.

\bibitem[\protect\citename{Takahashi \bgroup et al.\egroup }2018]{P18-1200}
Ryo Takahashi, Ran Tian, and Kentaro Inui.
\newblock 2018.
\newblock {Interpretable and Compositional Relation Learning by Joint Training
  with an Autoencoder}.
\newblock In {\em ACL}, pages 2148--2159.

\bibitem[\protect\citename{Tay \bgroup et al.\egroup
  }2017]{Tay:2017:RST:3018661.3018695}
Yi~Tay, Anh~Tuan Luu, Siu~Cheung Hui, and Falk Brauer.
\newblock 2017.
\newblock {Random Semantic Tensor Ensemble for Scalable Knowledge Graph Link
  Prediction}.
\newblock In {\em WSDM}, pages 751--760.

\bibitem[\protect\citename{Toutanova and Chen}2015]{toutanova-chen:2015:CVSC}
Kristina Toutanova and Danqi Chen.
\newblock 2015.
\newblock {Observed Versus Latent Features for Knowledge Base and Text
  Inference}.
\newblock In {\em CVSC}, pages 57--66.

\bibitem[\protect\citename{Toutanova \bgroup et al.\egroup
  }2016]{toutanova-EtAl:2016:P16-1}
Kristina Toutanova, Victoria Lin, Wen-tau Yih, Hoifung Poon, and Chris Quirk.
\newblock 2016.
\newblock {Compositional Learning of Embeddings for Relation Paths in Knowledge
  Base and Text}.
\newblock In {\em ACL}, pages 1434--1444.

\bibitem[\protect\citename{Trouillon \bgroup et al.\egroup
  }2016]{TrouillonWRGB16}
Th{\'{e}}o Trouillon, Johannes Welbl, Sebastian Riedel, {\'{E}}ric Gaussier,
  and Guillaume Bouchard.
\newblock 2016.
\newblock {Complex Embeddings for Simple Link Prediction}.
\newblock In {\em ICML}, pages 2071--2080.

\bibitem[\protect\citename{Tu \bgroup et al.\egroup }2017]{ijcai2017-399}
Cunchao Tu, Zhengyan Zhang, Zhiyuan Liu, and Maosong Sun.
\newblock 2017.
\newblock {TransNet: Translation-Based Network Representation Learning for
  Social Relation Extraction}.
\newblock In {\em IJCAI}, pages 2864--2870.

\bibitem[\protect\citename{{Vashishth} \bgroup et al.\egroup
  }2020]{interacte2020}
Shikhar {Vashishth}, Soumya {Sanyal}, Vikram {Nitin}, Nilesh {Agrawal}, and
  Partha {Talukdar}.
\newblock 2020.
\newblock {InteractE: Improving Convolution-based Knowledge Graph Embeddings by
  Increasing Feature Interactions}.
\newblock In {\em AAAI}.

\bibitem[\protect\citename{Veličković \bgroup et al.\egroup
  }2018]{velickovic2018graph}
Petar Veličković, Guillem Cucurull, Arantxa Casanova, Adriana Romero, Pietro
  Liò, and Yoshua Bengio.
\newblock 2018.
\newblock Graph attention networks.
\newblock In {\em ICLR}.

\bibitem[\protect\citename{Vu \bgroup et al.\egroup }2017]{NguyenECIR2017}
Thanh Vu, Dat~Quoc Nguyen, Mark Johnson, Dawei Song, and Alistair Willis.
\newblock 2017.
\newblock {Search Personalization with Embeddings}.
\newblock In {\em ECIR}, pages 598--604.

\bibitem[\protect\citename{Wang and Li}2016]{DBLP:conf/ijcai/WangL16}
Zhigang Wang and Juan{-}Zi Li.
\newblock 2016.
\newblock {Text-Enhanced Representation Learning for Knowledge Graph}.
\newblock In {\em IJCAI}, pages 1293--1299.

\bibitem[\protect\citename{Wang \bgroup et al.\egroup }2014]{AAAI148531}
Zhen Wang, Jianwen Zhang, Jianlin Feng, and Zheng Chen.
\newblock 2014.
\newblock {Knowledge Graph Embedding by Translating on Hyperplanes}.
\newblock In {\em AAAI}, pages 1112--1119.

\bibitem[\protect\citename{Wang \bgroup et al.\egroup
  }2016]{wang-EtAl:2016:P16-13}
Quan Wang, Jing Liu, Yuanfei Luo, Bin Wang, and Chin-Yew Lin.
\newblock 2016.
\newblock {Knowledge Base Completion via Coupled Path Ranking}.
\newblock In {\em ACL}, pages 1308--1318.

\bibitem[\protect\citename{{Wang} \bgroup et al.\egroup }2017]{8047276}
Q.~{Wang}, Z.~{Mao}, B.~{Wang}, and L.~{Guo}.
\newblock 2017.
\newblock Knowledge graph embedding: A survey of approaches and applications.
\newblock {\em IEEE Transactions on Knowledge and Data Engineering},
  29(12):2724--2743.

\bibitem[\protect\citename{Wei \bgroup et al.\egroup
  }2016]{wei-zhao-liu:2016:EMNLP2016}
Zhuoyu Wei, Jun Zhao, and Kang Liu.
\newblock 2016.
\newblock {Mining Inference Formulas by Goal-Directed Random Walks}.
\newblock In {\em EMNLP}, pages 1379--1388.

\bibitem[\protect\citename{West \bgroup et al.\egroup
  }2014]{West:2014:KBC:2566486.2568032}
Robert West, Evgeniy Gabrilovich, Kevin Murphy, Shaohua Sun, Rahul Gupta, and
  Dekang Lin.
\newblock 2014.
\newblock {Knowledge Base Completion via Search-based Question Answering}.
\newblock In {\em WWW}, pages 515--526.

\bibitem[\protect\citename{Weston and Bordes}2014]{westonembedding}
Jason Weston and Antoine Bordes.
\newblock 2014.
\newblock {Embedding Methods for NLP}.
\newblock In {\em EMNLP 2014 tutorial}.

\bibitem[\protect\citename{Xiao \bgroup et al.\egroup
  }2016]{xiao-huang-zhu:2016:P16-1}
Han Xiao, Minlie Huang, and Xiaoyan Zhu.
\newblock 2016.
\newblock {TransG : A Generative Model for Knowledge Graph Embedding}.
\newblock In {\em ACL}, pages 2316--2325.

\bibitem[\protect\citename{Xiao \bgroup et al.\egroup }2017]{0005HZ16}
Han Xiao, Minlie Huang, and Xiaoyan Zhu.
\newblock 2017.
\newblock {SSP:} semantic space projection for knowledge graph embedding with
  text descriptions.
\newblock In {\em AAAI}, pages 3104--3110.

\bibitem[\protect\citename{Xie \bgroup et al.\egroup }2017]{xie-EtAl:2017:Long}
Qizhe Xie, Xuezhe Ma, Zihang Dai, and Eduard Hovy.
\newblock 2017.
\newblock {An Interpretable Knowledge Transfer Model for Knowledge Base
  Completion}.
\newblock In {\em ACL}, pages 950--962.

\bibitem[\protect\citename{Xie \bgroup et al.\egroup
  }2020]{xie-etal-2020-reinceptione}
Zhiwen Xie, Guangyou Zhou, Jin Liu, and Jimmy~Xiangji Huang.
\newblock 2020.
\newblock {R}e{I}nception{E}: Relation-aware inception network with joint
  local-global structural information for knowledge graph embedding.
\newblock In {\em ACL}, pages 5929--5939.

\bibitem[\protect\citename{Xu and Li}2019]{xu-li-2019-relation}
Canran Xu and Ruijiang Li.
\newblock 2019.
\newblock {Relation Embedding with Dihedral Group in Knowledge Graph}.
\newblock In {\em ACL}, pages 263--272.

\bibitem[\protect\citename{Yang \bgroup et al.\egroup }2015]{yang-etal-2015}
Bishan Yang, Wen-tau Yih, Xiaodong He, Jianfeng Gao, and Li~Deng.
\newblock 2015.
\newblock {Embedding Entities and Relations for Learning and Inference in
  Knowledge Bases}.
\newblock In {\em ICLR}.

\bibitem[\protect\citename{Yang \bgroup et al.\egroup }2017]{NIPS2017_6826}
Fan Yang, Zhilin Yang, and William~W Cohen.
\newblock 2017.
\newblock {Differentiable Learning of Logical Rules for Knowledge Base
  Reasoning}.
\newblock In {\em NIPS}, pages 2316--2325.

\bibitem[\protect\citename{Yin \bgroup et al.\egroup
  }2018]{yin-etal-2018-recurrent}
Wenpeng Yin, Yadollah Yaghoobzadeh, and Hinrich Sch{\"u}tze.
\newblock 2018.
\newblock {Recurrent One-Hop Predictions for Reasoning over Knowledge Graphs}.
\newblock In {\em COLING}, pages 2369--2378.

\bibitem[\protect\citename{Yoon \bgroup et al.\egroup
  }2016]{yoon-EtAl:2016:N16-1}
Hee-Geun Yoon, Hyun-Je Song, Seong-Bae Park, and Se-Young Park.
\newblock 2016.
\newblock {A Translation-Based Knowledge Graph Embedding Preserving Logical
  Property of Relations}.
\newblock In {\em NAACL-HLT}, pages 907--916.

\bibitem[\protect\citename{Zhang \bgroup et al.\egroup
  }2016]{10.1145/2939672.2939673}
Fuzheng Zhang, Nicholas~Jing Yuan, Defu Lian, Xing Xie, and Wei-Ying Ma.
\newblock 2016.
\newblock Collaborative knowledge base embedding for recommender systems.
\newblock In {\em KDD}, page 353–362.

\bibitem[\protect\citename{Zhang \bgroup et al.\egroup }2017]{Zhang_2017_CVPR}
Hanwang Zhang, Zawlin Kyaw, Shih-Fu Chang, and Tat-Seng Chua.
\newblock 2017.
\newblock Visual translation embedding network for visual relation detection.
\newblock In {\em CVPR}, pages 5532--5540.

\bibitem[\protect\citename{Zhang \bgroup et al.\egroup
  }2018]{zhang-etal-2018-knowledge}
Zhao Zhang, Fuzhen Zhuang, Meng Qu, Fen Lin, and Qing He.
\newblock 2018.
\newblock Knowledge graph embedding with hierarchical relation structure.
\newblock In {\em EMNLP}, pages 3198--3207.

\bibitem[\protect\citename{Zhang \bgroup et al.\egroup }2019]{NIPS2019_8541}
Shuai Zhang, Yi~Tay, Lina Yao, and Qi~Liu.
\newblock 2019.
\newblock Quaternion knowledge graph embeddings.
\newblock In {\em Advances in Neural Information Processing Systems 32}, pages
  2735--2745.

\bibitem[\protect\citename{Zhang \bgroup et al.\egroup
  }2020]{Zhang2020Efficient}
Yuyu Zhang, Xinshi Chen, Yuan Yang, Arun Ramamurthy, Bo~Li, Yuan Qi, and
  Le~Song.
\newblock 2020.
\newblock Efficient probabilistic logic reasoning with graph neural networks.
\newblock In {\em ICLR}.

\end{thebibliography}

\newpage

\section*{Appendix}

\subsection*{Triple Classification---Task Description}

The triple classification task  was first introduced by \newcite{NIPS2013_5028}, and since then it has been used to evaluate various embedding models. The aim of this task is to predict whether a triple $(h, r, t)$ is correct or not.  
For classification,  a relation-specific threshold $\theta_r$ is set for each relation type $r$. 
If the plausibility score of an unseen test triple  $(h, r, t)$ is higher than $\theta_r$ then the triple will be classified as correct, otherwise incorrect. 
Following \newcite{NIPS2013_5028}, the relation-specific thresholds are determined by maximizing the micro-averaged accuracy, which is a per-triple average, on the validation set.

\begin{table}[!ht]
\centering
\setlength{\tabcolsep}{0.4em}
\begin{tabular}{l|lllll}
\hline
\bf Dataset & $\mid\mathcal{E}\mid$ & $\mid\mathcal{R}\mid$  & \multicolumn{3}{l}{\#Triples in train/valid/test} \\
\hline
FB13 \cite{NIPS2013_5028} & 75,043  & 13 &  316,232 &  5,908 & 23,733\\
WN11 \cite{NIPS2013_5028} &38,696 & 11 & 112,581 & 2,609 & 10,544 \\
\hline
\end{tabular}
\caption{Statistics of the benchmark datasets for triple classification. In both WN11 and
FB13, each validation and test set also contains the
same number of incorrect triples as the number of
correct triples.}
\label{tab:datasets1}
\end{table}

\subsection*{Triple Classification---Datasets}

Information  about  benchmark  datasets  for  the triple classification task is  given  in  Table \ref{tab:datasets1}.  FB13 and WN11 \cite{NIPS2013_5028} are derived from the large real-world KG Freebase \cite{Bollacker:2008} and  the large lexical KG WordNet \cite{Miller:1995:WLD:219717.219748}, respectively. Note that when creating the  FB13 and WN11 datasets,  \newcite{NIPS2013_5028} already filtered out triples from the test set if either or both of their head and tail entities also appear in the training set in a different relation type or order.

\begin{table}[!ht]
\centering
\resizebox{9cm}{!}{
\begin{tabular}{l|ll|l}
\hline
\bf Method &\bf W11 & \bf F13 & Avg.  \\
\hline
CTransR \citep{AAAI159571} & 85.7 & - & - \\
TransR \citep{AAAI159571} & 85.9 & 82.5 & 84.2\\
TransD \citep{ji-EtAl:2015:ACL-IJCNLP} & {86.4} & \textbf{89.1} & {87.8} \\
TEKE\_H  \citep{DBLP:conf/ijcai/WangL16}  & 84.8 & 84.2 & 84.5\\
TranSparse-S \citep{JiLH016} & {86.4} & 88.2 & 87.3\\
TranSparse-US \citep{JiLH016} & {86.8} & 87.5 & 87.2\\
ConvKB \citep{NguyenNNP2017} [*] & \underline{87.6} & {88.8} & \underline{88.2} \\
TransE-HRS \cite{zhang-etal-2018-knowledge} & 86.8 & 88.4 & 87.6\\
DISTMULT-HRS  \cite{zhang-etal-2018-knowledge} & \textbf{88.9} & \underline{89.0} & \textbf{89.0} \\
\hline
NTN \citep{NIPS2013_5028} & 70.6 & 87.2 & 78.9 \\
TransH \citep{AAAI148531} & 78.8 & 83.3 & 81.1 \\
SLogAn \citep{LiangF15} & 75.3 & 85.3 & 80.3 \\
KG2E \citep{He:2015} & 85.4 & 85.3 & 85.4  \\
Bilinear-\textsc{comp} \citep{guu-miller-liang:2015:EMNLP} & 77.6 & 86.1 & 81.9 \\
TransE-\textsc{comp} \citep{guu-miller-liang:2015:EMNLP} & 80.3 & 87.6 & 84.0 \\
TransR-FT \citep{FengHWZHZ16} & 86.6 & 82.9 & 84.8 \\
TransG \citep{xiao-huang-zhu:2016:P16-1} & {87.4} & 87.3 & 87.4 \\
lppTransD \citep{yoon-EtAl:2016:N16-1}  & 86.2 & {88.6} & 87.4 \\
TransE \citep{NIPS2013_5071} [*] & 86.5 & 87.5 & 87.0 \\
TransE-NMM \citep{NguyenCoNLL2016} & {86.8} & {88.6} & {87.7}\\
TranSparse-DT \citep{8057770} & {87.1} & 87.9 & {87.5}  \\

\hline
\end{tabular}
}
\caption{Accuracy results (in \%) for triple classification on WN11 (labeled as \textbf{W11}) and FB13 (labeled as \textbf{F13}) test sets, which are taken from the corresponding papers. ``Avg.'' denotes the averaged accuracy. [*] denotes that scores are taken from  \protect\newcite{SWJ318}.}
\label{tab:compared1}
\end{table}

\subsection*{Triple Classification---Main Results}

Table \ref{tab:compared1} presents the triple classification results  of KG completion models on the WN11 and FB13 datasets.  
The first 9 rows report the performance of models that use TransE/DISTMULT  to initialize the entity and relation vectors.  
The last 12 rows present the accuracy of models with randomly initialized parameters.  Note that there are higher triple classification results computed for NTN, Bilinear-\textsc{comp} and  TransE-\textsc{comp} when entity vectors are initialized by averaging the pre-trained GloVe word vectors  \citep{Pennington14}.
 It  is not surprising because many entity names in WordNet and Freebase are lexically meaningful. However, this is not always the case w.r.t. many domain-specific KGs.

\end{document}